\documentclass{article}
\usepackage{iclr2027_conference,times}

\usepackage{amsmath,amssymb,amsfonts,amsthm,mathtools}

\usepackage[utf8]{inputenc}
\usepackage[T1]{fontenc}
\usepackage{hyperref}
\usepackage{url}
\usepackage{booktabs}
\usepackage{tabularx}
\usepackage{nicefrac}
\usepackage{microtype}
\usepackage{xcolor}
\usepackage{colortbl}
\usepackage{graphicx}
\usepackage{wrapfig}
\usepackage{needspace}
\usepackage{subfig}
\usepackage{multirow}
\usepackage{algorithm}
\usepackage{algorithmic}

\definecolor{MetaPairBg}{HTML}{F2ECE3}
\definecolor{MetaPairAccent}{HTML}{865D3D}
\definecolor{MetaPairText}{HTML}{3F352D}
\definecolor{MetaRelBg}{HTML}{EAF2F8}
\definecolor{MetaRelText}{HTML}{2F5F87}
\definecolor{MetaAbsBg}{HTML}{EAF5EA}
\definecolor{MetaAbsText}{HTML}{2F7D32}

\usepackage[capitalize,noabbrev]{cleveref}

\theoremstyle{plain}

\theoremstyle{definition}

\theoremstyle{remark}

\crefname{assumption}{Assumption}{Assumptions}
\Crefname{assumption}{Assumption}{Assumptions}

\newcommand{\R}{\mathbb{R}}

\newcommand{\E}{\mathbb{E}}

\newcommand{\Vocab}{\mathcal{V}}
\newcommand{\Dtrain}{\mathcal{D}_{\mathrm{train}}}

\DeclareMathOperator{\KL}{D_{\mathrm{KL}}}

\title{MetaOPD: Meta-Learned Token Weighting for On-Policy Distillation}

\author{Zipeng Wang$^{1}$, 
Xinpeng Dong$^{2}$, 
Yuefan Wang$^{2}$, 
Pingchen Lu$^{3}$, 
Xian Wei$^{1}$, 
Kun Kuang$^{2}$, \\
\textbf{Fei Wu}$^{2}$, 
\textbf{Zhongxiang Dai}$^{3}$\thanks{Corresponding authors}\hspace{1.6mm}, 
\textbf{Min Zhang}$^{2*}$\\
$^{1}$East China Normal University,
$^{2}$Zhejiang University, \\
$^{3}$The Chinese University of Hong Kong, Shenzhen
}

\iclrfinalcopy

\begin{document}

\maketitle

\begin{abstract}

On-policy distillation (OPD) trains a student on its own generated responses using token-level teacher supervision. However, uniform weighting overlooks differences in token learning value, while existing weighting methods rely on predefined mappings from prediction signals to token weights. These mappings are not learned from the effectiveness of the resulting student updates, limiting their ability to adapt to evolving learning needs. In this paper, we propose \textbf{MetaOPD}, a bilevel optimization framework that jointly learns the student model and a lightweight token-weighting network. The inner objective updates the student through weighted OPD, while the outer objective optimizes the weighting network using validation loss on reference solutions after a virtual student update. Differentiating through this update connects weighting decisions to their effects on post-update performance, allowing the mapping from prediction signals to token weights to evolve alongside the student. Experiments on six mathematical reasoning and three out-of-domain datasets, covering two student scales and seven baselines, demonstrate the effectiveness of MetaOPD, with Avg@8/Pass@8 gains over OPD of 1.99/5.97 percentage points for the 0.6B student and 2.25/6.41 points for the 1.7B student.
\end{abstract}


%



\section{Introduction}

Off-policy distillation trains a student on teacher-generated sequences, which can lead to a distribution mismatch between training and inference~\citep{lin2020autoregressive,agarwal2024policy}. To mitigate this issue, on-policy distillation (OPD) trains the student on its own generated trajectories, with a frozen teacher providing dense token-level supervision. OPD typically employs a mode-seeking reverse-KL objective that encourages the student to capture the teacher’s high-probability behaviors~\citep{wen2023f,gu2024minillm}. However, experiments by~\citep{li2026rethinking} show that tokens differ substantially in their learning value, a distinction that standard OPD’s uniform weighting fails to capture.

Recent approaches address this limitation by allocating token-level weights through \textbf{static proxy rules}. For example, EOPD~\citep{jin2026entropy} uses predictive entropy, while TIP~\citep{xu2026tip} leverages teacher--student disagreement. These methods provide useful signals for allocating supervision based on predictive uncertainty or differences between model predictions. However, their mappings from proxy signals to token weights are predefined rather than adapted to the student's actual learning gains. As training progresses, the student's capabilities and generated contexts evolve, and the learning benefit associated with the same proxy signal may change accordingly. Thus, even when token weights vary across inputs, a fixed weighting rule may not transfer effectively across training stages or task contexts. \textit{The central challenge is to learn a weighting rule that adapts supervision to the student's evolving learning needs through feedback from the resulting updates.}

\begin{figure}[t]
    \centering
    \captionsetup{skip=4pt}
    \captionsetup[subfloat]{farskip=0pt,nearskip=0pt}
    \subfloat{\label{fig:intro-motivation-a}%
        \includegraphics[width=0.560606\linewidth,trim=0bp 0bp 174bp 0bp,clip]{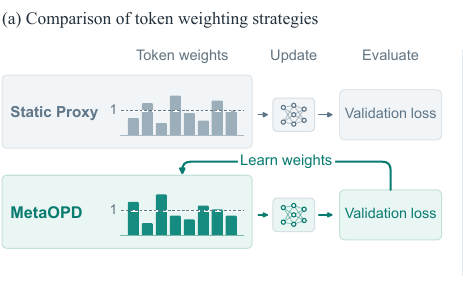}%
    }%
    \subfloat{\label{fig:intro-motivation-b}%
        \includegraphics[width=0.439394\linewidth,trim=222bp 0bp 0bp 0bp,clip]{figure1_update_feedback.pdf}%
    }
    \vspace{-1mm}
    \caption{\textbf{Motivating experiment on token weighting strategies.} (a) Static proxy methods assign token weights using predefined signals, while MetaOPD learns weights through validation-loss feedback. (b) Validation loss reductions relative to Uniform OPD for MetaOPD, EOPD~\citep{jin2026entropy}, TIP~\citep{xu2026tip}, and random token weighting. MetaOPD achieves the largest reduction, supporting the benefit of learning token weights aligned with the validation objective.}
    \vspace{-3.5mm}
    \label{fig:intro-motivation}
\end{figure}

To examine whether different token weighting strategies translate into actual learning gains, we conduct a motivating experiment. As illustrated in~\cref{fig:intro-motivation-a}, we compare assigning weights through predefined proxies with learning weights. Figure~1(b) reports validation loss reductions relative to Uniform OPD, with larger values indicating greater gains. EOPD yields only a modest improvement, while TIP and random weighting perform close to the uniform baseline. In contrast, learning weights through validation-loss feedback produces a substantially larger improvement. These results suggest that, in this setting, predefined rules do not fully exploit the potential benefits of token weighting, whereas the student's update performance provides valuable feedback for weight allocation. The key is therefore not only to distinguish tokens using prediction signals, but also to adapt the signal-to-weight mapping based on the effectiveness of the resulting updates. This leads to our central question: \textbf{How can we use the student's post-update performance to learn a token weighting rule that translates prediction signals into effective supervision?}


In this paper, we take a step forward by proposing \textbf{MetaOPD}, an on-policy distillation framework that jointly optimizes the student model and the token-weighting rule. Specifically, we introduce a lightweight token-weighting network that maps prediction signals to token weights. The student and the weighting network perform two coupled learning tasks: \textit{the student improves through weighted teacher supervision}, while \textit{the weighting network learns to allocate supervision based on the effectiveness of student updates}. To achieve this, we formulate their joint learning as a bilevel optimization problem. At the inner level, we update the student on its own generated responses using an OPD loss weighted by the network's outputs. At the outer level, we evaluate the student's loss on validation reference solutions after a virtual update and use this loss to optimize the weighting network. Differentiating through the virtual update allows the student's post-update performance to guide token weight allocation, enabling the network to learn which allocations lead to greater subsequent improvements. Alternating between the two levels allows the student and the weighting rule to evolve together, replacing predefined proxy weighting with adaptive supervision allocation driven by update outcomes. Our main contributions are summarized as follows:
\begin{itemize}
    \item We identify a gap between static proxy weighting and actual learning gains in OPD, and empirically motivate learning token weights from post-update performance feedback.
    \item We propose MetaOPD, a bilevel framework that jointly optimizes the student through weighted OPD and a lightweight token-weighting network through update validation loss.
    \item We conduct extensive experiments across six reasoning and three out-of-domain datasets on two student scales against seven baselines, demonstrating the effectiveness of our MetaOPD.
\end{itemize}

\section{Related Work}

\textbf{On-policy distillation (OPD).}
On-policy distillation (OPD) trains a student to match teacher predictions on student-generated trajectories~\citep{agarwal2024policy}.
Compared with distillation on fixed data or teacher-generated sequences~\citep{hinton2015distilling,kim2016sequence}, it reduces training--inference mismatch by providing teacher feedback at prefixes the student actually visits.
One line of work improves OPD through the design of distillation objectives and optimization strategies~\citep{gu2024minillm,ko2024distillm,yang2026learning,yu2026mismatch}.
Another focuses on selecting training prompts, improving rollout trajectories, and refining teacher supervision~\citep{li2026rethinking,xu2026pass,qu2026spot}.
Since tokens differ in learning value, assigning different weights to tokens within the OPD loss offers another way to improve OPD.

\textbf{Static proxy weighting OPD.}
Static proxy weighting OPD methods start from the observation that dense teacher feedback is not equally useful at every token.
Existing approaches use signals such as entropy and teacher--student disagreement to identify informative tokens and emphasize their supervision through token selection, loss adjustment, or reweighting~\citep{jin2026entropy,ko2026scaling,xu2026tip,wang2026not,li2026filter}.
Another line accounts for changes in supervision reliability along a trajectory, using accumulated prefix discrepancy or token position to guide weighting~\citep{xie2026position,liu2026teacher}.
However, these methods use predefined mappings from proxy signals to supervision allocation, without learning from post-update feedback how those mappings should adapt to the student's evolving learning needs.

\textbf{Bilevel Optimization.}
Bilevel optimization provides a framework for learning example or token weights using feedback on the updated model.
Early meta-reweighting methods used this structure to learn example importance under class imbalance and label noise~\citep{ren2018learning,shu2019meta}. 
Recent studies have extended this principle to language-model training, using outer-loop feedback to replace fixed weighting rules with update-aware supervision~\citep{hu2023meta,pan2025scalebio,yu2025llm,zhang2025dorm,li2026learning}. 
Across these settings, a common view emerges: the value of a training signal is not fixed, but depends on the model update it induces and the reference objective used to assess that update. 
MetaOPD applies this principle to token weighting within the OPD loss on student-generated trajectories.

\section{Methodology}
\label{sec:method}

This section first introduces the token-weighted on-policy distillation objective (\S\ref{sec:token-weighted-opd}), then presents the bilevel optimization procedure of MetaOPD (\S\ref{sec:bilevel-reweighting}), and finally explains the role of individual tokens in the weighting-network update and describes the gradient computation (\S\ref{sec:token-credit}).

\subsection{Token-Weighted On-Policy Distillation}
\label{sec:token-weighted-opd}
Let \(\Dtrain\) denote the training distribution over question--solution pairs.
We write \(\Dtrain^X\) for the corresponding distribution of questions. The student \(\pi_\theta\) has trainable parameters \(\theta\), and \(\pi_{\mathrm T}\) is the frozen teacher. At the start of each rollout round, we set \(\theta_{\mathrm{old}}\leftarrow\theta\). For \(x_i\sim\Dtrain^X\), we sample a response from \(\pi_{\theta_{\mathrm{old}}}\):
\begin{equation}
    y_i=(a_{i,1},\ldots,a_{i,T_i})
    \sim\pi_{\theta_{\mathrm{old}}}(\cdot\mid x_i),
    \qquad s_{i,t}=(x_i,a_{i,<t}),
    \label{eq:old-policy-rollout}
\end{equation}
where \(s_{i,t}\) is the visited prefix and \(a_{i,t}\) the sampled token. The mask \(m_{i,t}\in\{0,1\}\) marks valid completion positions, including an observed end-of-sequence token.

\begin{figure}[t]
    \centering
    \includegraphics[width=\linewidth]{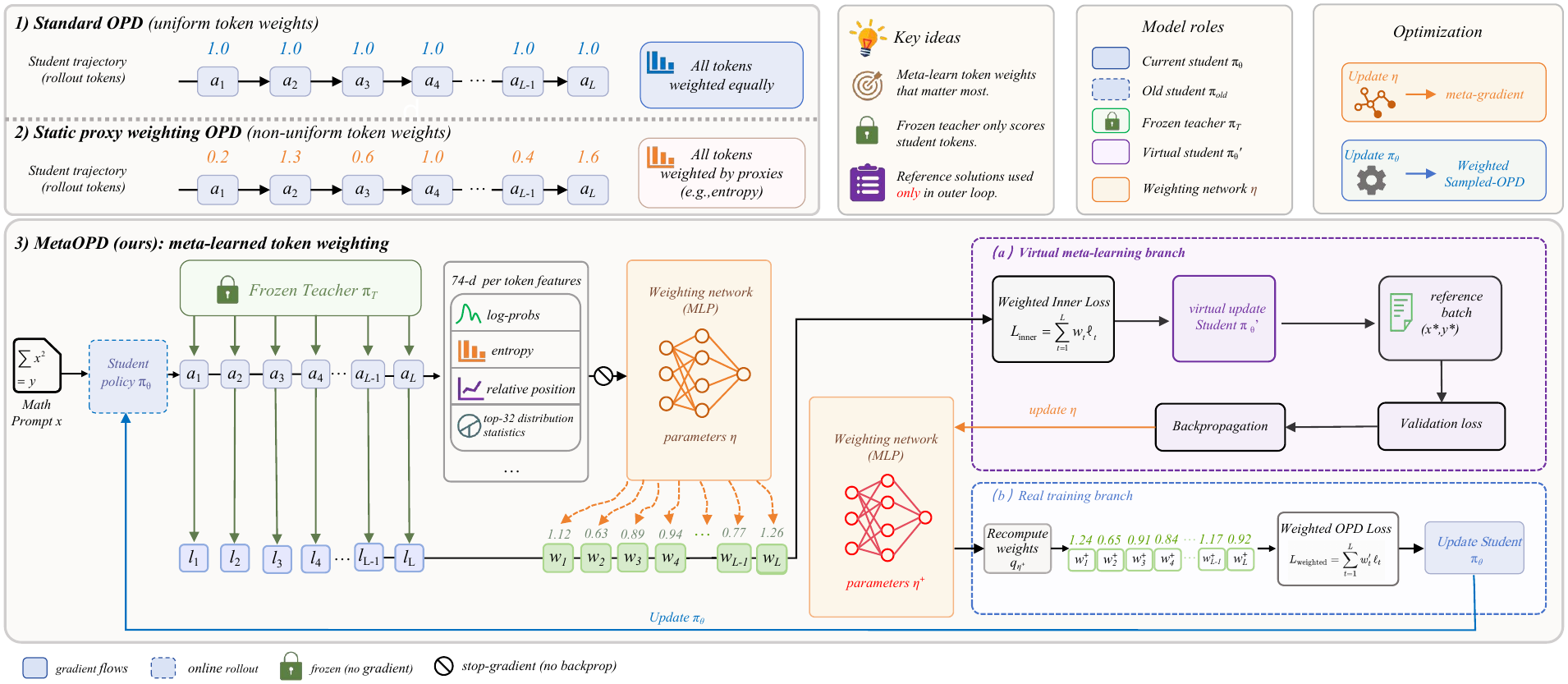}
    \vspace{-3mm}
    \caption{\textbf{Overview of MetaOPD.}
    (1) Standard OPD uses uniform token weights, while (2) static proxy weighting OPD assigns token weights through predefined rules based on proxy signals.
    (3) MetaOPD uses two branches: (a) the weighting network learns from the student's validation loss after a virtual weighted OPD update; (b) the updated network recomputes weights for the actual student update. The schematic sums omit valid-token normalization, and weights in the actual-update branch refer to those recomputed after the weighting-network update.}
    \vspace{-3mm}
    \label{fig:metaopd-method}
\end{figure}

For each sampled token, write \(u^\theta_{i,t}=\log\pi_\theta(a_{i,t}\mid s_{i,t})\), with \(u^{\mathrm{old}}_{i,t}\) denoting the rollout-time log-probability and \(u^{\mathrm T}_{i,t}\) the teacher log-probability after aligning token identities. All three use full-vocabulary normalization; Appendix~\ref{app:sampled-opd-math} specifies the teacher-view mapping. The detached teacher-minus-rollout log-probability difference and the current-to-old probability ratio are
\begin{equation}
    A_{i,t}=\operatorname{sg}[u^{\mathrm T}_{i,t}-u^{\mathrm{old}}_{i,t}],
    \qquad r_{i,t}(\theta)=\exp(u^\theta_{i,t}-u^{\mathrm{old}}_{i,t}),
    \label{eq:sampled-opd-state}
\end{equation}
where \(\operatorname{sg}\) preserves its argument's value but blocks gradients through it. Following the sampled reverse-KL update used in Relay-OPD~\citep{xu2026pass}, let \(r^{\mathrm{clip}}_{i,t}=\operatorname{clip}(r_{i,t},1-\epsilon_l,1+\epsilon_h)\), with clipping margins \(\epsilon_l,\epsilon_h\), and define the per-token OPD term as
\begin{equation}
    \ell^{\mathrm{OPD}}_{i,t}(\theta)
    =\max\!\left(-A_{i,t}r_{i,t}(\theta),-A_{i,t}r^{\mathrm{clip}}_{i,t}\right).
    \label{eq:sampled-opd-token-loss}
\end{equation}
At \(\theta=\theta_{\mathrm{old}}\), the unclipped form gives an unbiased single-action estimate of the student-to-teacher reverse-KL gradient at the fixed visited prefix, using the induced teacher distribution defined in Appendix~\ref{app:sampled-opd-math}. Away from that policy, the cached log-probability difference and clipping define a local surrogate.

A shared weighting network \(q_\eta\), with trainable parameters \(\eta\), maps each token's descriptor \(f_{i,t}\in\R^{74}\) to a scalar score. The descriptor contains student and teacher log-probabilities of the sampled token, entropy, prediction agreement, position, and ranked top-32 log-probability profiles (\cref{tab:descriptor}). These profiles are inputs only and do not restrict the OPD loss vocabulary. The feature \(\delta_{i,t}=u^{\mathrm T}_{i,t}-u^\theta_{i,t}\) uses current-student log-probabilities, whereas \(A_{i,t}\) uses cached rollout-policy log-probabilities. All numeric inputs are detached; descriptors are constructed without reference solutions.

We convert the network scores into positive weights with unit mean within each response. For \(n_i=\sum_t m_{i,t}>0\), define
\begin{equation}
\begin{aligned}
    z_{i,t}&=q_\eta(\operatorname{sg}[f_{i,t}]),
    &\bar z_i&=\frac{1}{n_i}\sum_s m_{i,s}z_{i,s},\\
    \widetilde w_{\eta,i,t}&=1+\rho\tanh(z_{i,t}-\bar z_i),
    &w_{\eta,i,t}&=m_{i,t}\widetilde w_{\eta,i,t}
       \frac{n_i}{\sum_s m_{i,s}\widetilde w_{\eta,i,s}}.
\end{aligned}
\label{eq:weight-normalization}
\end{equation}
Centering expresses each token's score relative to the mean over valid tokens in the same response. Normalization then keeps the mean weight over these tokens at one, allowing the network to adjust their relative contributions within the OPD loss. Both operations are differentiated with respect to \(\eta\). We use residual scale \(\rho=0.25\) and a zero-initialized output head, so all valid weights start at one. Completely masked responses receive zero weights. The architecture, weight bounds, and numerical safeguards appear in Appendices~\ref{app:weight-normalization-math} and~\ref{app:reproducibility}.

For an optimizer minibatch \(B\), the weighted OPD objective is
\begin{equation}
    \mathcal L_{\mathrm{in}}(\theta,\eta;B)
    =\frac{1}{N_B}\sum_{i,t}
      m_{i,t}w_{\eta,i,t}\ell^{\mathrm{OPD}}_{i,t}(\theta),
    \qquad N_B=\sum_{i,t}m_{i,t}.
    \label{eq:inner-loss}
\end{equation}
The numerator and valid-token count are reduced over the complete optimizer minibatch. Setting all valid weights to one recovers uniform sampled OPD.

\subsection{Bilevel Optimization of MetaOPD}
\label{sec:bilevel-reweighting}

\Cref{fig:metaopd-method} illustrates the bilevel procedure: the student learns from weighted OPD, while the weighting network learns from the student's validation loss after a virtual student update. Here validation loss is the teacher-forced token loss on training reference solutions, used as the outer objective defined below. Following gradient-based meta-reweighting and Meta-Weight-Net~\citep{ren2018learning,shu2019meta}, each optimizer minibatch involves a virtual student update, a weighting-network update, and an actual student update.

\textbf{Virtual student update.}
Teacher statistics are cached with the rollout. At the start of each optimizer minibatch, we compute current-student statistics and detach the descriptors used to obtain the token weights. Let \(\mathsf S\) collect the incoming student AdamW moments, step counters, and parameter-group hyperparameters. Write \(\boldsymbol g_{\mathrm{in}}(\eta):=\nabla_\theta\mathcal L_{\mathrm{in}}(\theta,\eta;B)\) for the student gradient, whose dependence on \(\eta\) is through the token weights. One virtual step defines the virtual student parameters
\begin{equation}
    \theta'(\eta)=\mathcal A_{\mathrm{AdamW}}\!\left(
       \theta,\mathcal C\!\left(\boldsymbol g_{\mathrm{in}}(\eta)\right);\mathsf S\right).
    \label{eq:virtual-update}
\end{equation}
The virtual and actual student steps use the same AdamW update rule~\citep{loshchilov2017decoupled}. Here \(\mathcal C\) applies full-model global-norm clipping, and \(\mathcal A_{\mathrm{AdamW}}\) reads the current student optimizer state and scheduled learning rates. Differentiation with respect to \(\eta\) holds \(\theta\), \(\mathsf S\), the rollout, and detached descriptors fixed. The dependence through the token weights and student gradient is retained, including the new virtual moments, clipping coefficient, and AdamW parameter map. This operation changes neither the real parameters nor the stored moments, step counters, or scheduler. Appendix~\ref{app:virtual-adamw} expands the two maps.

\textbf{Weighting-network update.}
Draw a batch \(B^{\mathrm{ref}}=\{(x_j^\star,y_j^\star)\}\) of prompts and reference solutions from a separately shuffled stream over \(\Dtrain\); overlap with the rollout stream is allowed. At \(\theta'(\eta)\), the student predicts each solution token given the prompt and preceding solution tokens. We compute validation loss under teacher forcing, averaging over valid target tokens:
\begin{equation}
    \mathcal L_{\mathrm{out}}(\theta';B^{\mathrm{ref}})
    =-\frac{1}{N_{\mathrm{ref}}}\sum_{j,t}m^{\mathrm{ref}}_{j,t}
       \log\pi_{\theta'}(y_{j,t}^\star\mid x_j^\star,y_{j,<t}^\star),
    \qquad N_{\mathrm{ref}}=\sum_{j,t}m^{\mathrm{ref}}_{j,t}.
    \label{eq:outer-loss}
\end{equation}
The mask \(m^{\mathrm{ref}}_{j,t}\) selects valid reference-solution tokens; \(N_{\mathrm{ref}}\) is counted separately from \(N_B\).
To regularize departures from uniform weighting, define
\begin{equation}
    \mathcal R_w(\eta;B)=\frac{1}{N_B}\sum_{i,t}
      m_{i,t}(w_{\eta,i,t}-1)^2.
    \label{eq:weight-regularizer}
\end{equation}
At global student step \(r\), let \(\lambda_w^{(r)}\) be the weight-regularization coefficient, fixed for this step. We optimize the one-step meta-objective
\begin{equation}
    \mathcal J_{\mathrm{meta}}(\eta)
    =\mathcal L_{\mathrm{out}}(\theta'(\eta);B^{\mathrm{ref}})
      +\lambda_w^{(r)}\mathcal R_w(\eta;B).
    \label{eq:meta-objective}
\end{equation}
Applying the chain rule to the meta-objective gives the \emph{meta-gradient} with respect to \(\eta\):
\begin{equation}
\begin{aligned}
    \nabla_\eta\mathcal J_{\mathrm{meta}}
    &=\left(\frac{\partial\theta'(\eta)}{\partial\eta}\right)^\top
      \nabla_{\theta'}\mathcal L_{\mathrm{out}}
      \bigl(\theta'(\eta);B^{\mathrm{ref}}\bigr)
      +\lambda_w^{(r)}\nabla_\eta\mathcal R_w(\eta;B).
\end{aligned}
\label{eq:bilevel-meta-gradient}
\end{equation}
The first term propagates the validation-loss gradient through the virtual update; the second comes from the weight regularizer. Let \(\mathsf M\) denote the weighting network's AdamW state and \(\operatorname{AdamWStep}\) an update returning parameters and state. One update per student minibatch gives
\begin{equation}
    (\eta^+,\mathsf M^+)
    =\operatorname{AdamWStep}_{\eta}
      (\eta,\nabla_\eta\mathcal J_{\mathrm{meta}};\mathsf M).
    \label{eq:meta-update}
\end{equation}
Section~\ref{sec:token-credit} expands this gradient into token contributions and describes its computation. The weighting-network learning rate and \(\lambda_w^{(r)}\) follow the student learning-rate schedule in Appendix~\ref{app:meta-schedule}.

\textbf{Actual student update.}
We discard the virtual parameters and recompute token weights with \(q_{\eta^+}\) on the same detached descriptors. The actual student gradient uses these weights as constants:
\begin{equation}
\begin{aligned}
    \boldsymbol g_{\mathrm{real}}
    &=\frac{1}{N_B}\sum_{i,t}m_{i,t}
      \operatorname{sg}[w_{\eta^+,i,t}]
      \nabla_\theta\ell^{\mathrm{OPD}}_{i,t}(\theta),\\
    (\theta^+,\mathsf S^+)
    &=\operatorname{AdamWStep}_{\theta}
      (\theta,\mathcal C(\boldsymbol g_{\mathrm{real}});\mathsf S).
\end{aligned}
\label{eq:actual-student-update}
\end{equation}
This step starts from the original \((\theta,\mathsf S)\), not from \(\theta'\), and advances the real optimizer state. 
Its objective contains only weighted OPD supervision on student-generated responses; validation loss trains the weighting network through the meta-objective.

\begin{algorithm}[H]
\caption{One MetaOPD rollout round}
\label{alg:metaopd}
\begin{algorithmic}[1]
\REQUIRE Student parameters and AdamW state \((\theta,\mathsf S)\); weighting parameters and meta-AdamW state \((\eta,\mathsf M)\); frozen teacher \(\pi_{\mathrm T}\)
\REQUIRE Separately shuffled reference-solution stream \(\mathcal G\); minibatch count \(K\); global student-step index \(r\)
\ENSURE Updated \((\theta,\mathsf S)\), \((\eta,\mathsf M)\), and \(r\)
\STATE Set \(\theta_{\mathrm{old}}\leftarrow\theta\); sample prompts from \(\Dtrain^X\) and generate student responses.
\STATE Query the teacher at visited prefixes; cache teacher statistics, old-policy log-probabilities, and masks.
\STATE Partition the rollout into \(K\) optimizer minibatches.
\FOR{each minibatch \(B\)}
    \STATE Draw \(B^{\mathrm{ref}}\leftarrow\operatorname{Next}(\mathcal G)\); compute and detach current-student descriptors.
    \STATE Compute token weights and \(\mathcal L_{\mathrm{in}}\) using \cref{eq:weight-normalization,eq:inner-loss}.
    \STATE \textbf{Virtual update:} construct \(\theta'(\eta)\) using \cref{eq:virtual-update}, leaving \((\theta,\mathsf S)\) unchanged.
    \STATE \textbf{Meta-update:} compute \(\nabla_\eta\mathcal J_{\mathrm{meta}}\) using \cref{eq:bilevel-meta-gradient}; obtain \((\eta^+,\mathsf M^+)\) using \cref{eq:meta-update}.
    \STATE Discard \(\theta'\); recompute weights with \(q_{\eta^+}\) on the same descriptors and detach them.
    \STATE \textbf{Student update:} obtain \((\theta^+,\mathsf S^+)\) from the original \((\theta,\mathsf S)\) using \cref{eq:actual-student-update}.
    \STATE Set \((\theta,\mathsf S)\leftarrow(\theta^+,\mathsf S^+)\), \((\eta,\mathsf M)\leftarrow(\eta^+,\mathsf M^+)\), and \(r\leftarrow r+1\).
\ENDFOR
\end{algorithmic}
\end{algorithm}

\subsection{Token-Level Interpretation and Gradient Computation}
\label{sec:token-credit}

We examine how a small change in a token's weight within the OPD loss affects validation loss after the virtual student update. Throughout, \(\nabla_\eta\mathcal L_{\mathrm{out}}\) denotes the gradient with respect to \(\eta\) obtained by differentiating \(\mathcal L_{\mathrm{out}}(\theta'(\eta);B^{\mathrm{ref}})\) through the virtual update, with the inputs fixed as in \S\ref{sec:bilevel-reweighting}.
Define the token gradient \(\boldsymbol g_{i,t}=\nabla_\theta\ell^{\mathrm{OPD}}_{i,t}(\theta)\). The detached descriptors make weights constant with respect to \(\theta\), while preserving their dependence on \(\eta\), so
\begin{equation}
    \boldsymbol g_{\mathrm{in}}(\eta)
    =\frac{1}{N_B}\sum_{i,t}m_{i,t}w_{\eta,i,t}\boldsymbol g_{i,t}.
    \label{eq:inner-gradient-dependence}
\end{equation}

Let \(\boldsymbol g_{\mathrm{out}}=\nabla_{\theta'}\mathcal L_{\mathrm{out}}(\theta';B^{\mathrm{ref}})\), evaluated at \(\theta'=\theta'(\eta)\).
Write \(J_{\mathcal C}\) for the clipping Jacobian at \(\boldsymbol g_{\mathrm{in}}(\eta)\), and \(J_{\mathcal A}\) for the AdamW parameter-map Jacobian with respect to its gradient input, evaluated at \(\mathcal C(\boldsymbol g_{\mathrm{in}}(\eta))\). Propagating the outer gradient through these maps defines
\begin{equation}
    \boldsymbol d=-J_{\mathcal C}^{\top}J_{\mathcal A}^{\top}\boldsymbol g_{\mathrm{out}}.
    \label{eq:optimizer-aware-direction}
\end{equation}
This derivative-based vector, rather than the actual student parameter change, determines each token's contribution to the weighting-network gradient:
\begin{equation}
    c_{i,t}=\langle\boldsymbol d,\boldsymbol g_{i,t}\rangle,
    \qquad
    \nabla_\eta\mathcal L_{\mathrm{out}}
    =-\frac{1}{N_B}\sum_{i,t}m_{i,t}c_{i,t}\nabla_\eta w_{\eta,i,t}.
    \label{eq:weight-hypergradient}
\end{equation}

For a valid token, \(\partial\mathcal L_{\mathrm{out}}/\partial w_{i,t}=-c_{i,t}/N_B\) when the other normalized-weight coordinates are held fixed. Thus, \(c_{i,t}>0\) means that a small increase in this weight alone locally reduces validation loss.
Learning rates are already included in \(J_{\mathcal A}\); adding \(\lambda_w^{(r)}\nabla_\eta\mathcal R_w\) to \cref{eq:weight-hypergradient} gives the complete meta-gradient in \cref{eq:bilevel-meta-gradient}. Appendix~\ref{app:bilevel-math} provides the intermediate Jacobian and chain-rule steps.

Response normalization couples the weight coordinates. Suppressing \(\eta\) for clarity, let \(\overline w_i=n_i^{-1}\sum_t m_{i,t}\widetilde w_{i,t}\). Differentiating \(w_{i,t}=\widetilde w_{i,t}/\overline w_i\) on valid positions gives
\begin{equation}
    \frac{\partial\mathcal L_{\mathrm{out}}}{\partial\widetilde w_{i,j}}
    =-\frac{c_{i,j}-\bar c_i^{\,w}}{N_B\overline w_i},
    \qquad
    \bar c_i^{\,w}=\frac{1}{n_i}\sum_t m_{i,t}w_{i,t}c_{i,t}.
    \label{eq:relative-token-credit}
\end{equation}
The gradient with respect to a raw weight therefore depends on \(c_{i,j}-\bar c_i^{\,w}\), the difference from the response's weighted mean. Score centering, the shared network, and regularization further affect the update, so \(c_{i,j}>0\) need not lead to an increase in the learned weight. This derivative describes a local effect; it does not guarantee finite-step or long-term improvement. Appendix~\ref{app:weight-normalization-math} expands the normalization derivative.

Reverse-mode vector--Jacobian products (VJPs) first compute \(\boldsymbol g_{\mathrm{out}}\), then propagate it through AdamW and clipping to obtain \(\boldsymbol d\). A Jacobian--vector product (JVP) of the vector of per-token OPD terms at the current \(\theta\), in direction \(\boldsymbol d\), gives all \(c_{i,t}\) values without materializing the token-by-parameter Jacobian. Recomputing weights with their \(\eta\)-dependence intact, we differentiate the following objective while treating \(c_{i,t}\) as constants:
\begin{equation}
    \widehat{\mathcal L}_{\mathrm{meta}}^{(r)}(\eta;B,\boldsymbol c)
    =-\frac{1}{N_B}\sum_{i,t}m_{i,t}
      w_{\eta,i,t}\operatorname{sg}[c_{i,t}]
      +\lambda_w^{(r)}\mathcal R_w(\eta;B).
    \label{eq:compatibility-surrogate}
\end{equation}
Holding \(c_{i,t}\) fixed retains the Jacobian-product values required by the chain rule without differentiating them again. The resulting \(\eta\)-gradient equals the complete gradient in \cref{eq:bilevel-meta-gradient} at the iterate where these values were computed; this is a first-derivative equality at that point, not a global identity between objectives. Appendix~\ref{app:bilevel-math} verifies the equality.

\begin{table}[t]
\centering
\footnotesize
\setlength{\tabcolsep}{2.0pt}
\renewcommand{\arraystretch}{1.12}
\caption{Mathematical-reasoning results (\%) with a Qwen3-8B teacher. Avg. is the unweighted average over six benchmarks; benchmark and Avg. scores are rounded separately. Abs. \(\Delta_{\mathrm{OPD}}\) reports MetaOPD--OPD in percentage points. Rel. \(\Delta_{\mathrm{OPD}}\) reports the same difference as a percentage of the OPD score. Delta values are calculated from the displayed scores. Bold marks the best result in each student setting, including ties.}
\label{tab:main-results}
\resizebox{\textwidth}{!}{%
\begin{tabular}{@{}l*{14}{c}@{}}
\toprule
\multirow{2}{*}{Method}
& \multicolumn{7}{c}{Avg@8}
& \multicolumn{7}{c}{Pass@8} \\
\cmidrule(lr){2-8}
\cmidrule(l){9-15}
& MATH500 & AMC23 & Minerva & HMMT & AIME24 & AIME25 & \textbf{Avg.}
& MATH500 & AMC23 & Minerva & HMMT & AIME24 & AIME25 & \textbf{Avg.} \\
\midrule
\rowcolor{MetaPairBg}
\multicolumn{15}{c}{%
    \textcolor{MetaPairAccent}{\bfseries\textsc{Teacher}}\;\textcolor{MetaPairText}{\bfseries Qwen3-8B}
    \quad\textcolor{MetaPairAccent}{\(\rightsquigarrow\)}\quad
    \textcolor{MetaPairAccent}{\bfseries\textsc{Student}}\;\textcolor{MetaPairText}{\bfseries Qwen3-0.6B-Base}
} \\
\midrule
KD       & 47.90 & 23.75 & 14.25 & 0.42 & 2.50 & 0.83 & 14.94 & 69.20 & 50.00 & 31.25 & 3.33 & 6.67 & 6.67 & 27.85 \\
GRPO     & \textbf{53.20} & 28.44 & 16.31 & 0.83 & \textbf{4.17} & 1.25 & 17.37 & 75.00 & 57.50 & 32.35 & 6.67 & 6.67 & 10.00 & 31.36 \\
OPD      & 50.28 & 25.00 & 15.95 & 0.42 & 2.92 & 1.25 & 15.97 & 73.00 & 55.00 & 31.62 & 3.33 & 10.00 & 6.67 & 29.94 \\
EOPD     & 50.40 & 27.50 & 16.08 & 1.67 & 3.75 & 1.25 & 16.78 & 75.60 & 52.50 & 34.19 & 6.67 & 13.33 & 6.67 & 31.49 \\
FiRe-OPD & 50.98 & 29.38 & 16.27 & 0.83 & 2.92 & 0.83 & 16.87 & 73.60 & \textbf{62.50} & 31.99 & 3.33 & 10.00 & 6.67 & 31.35 \\
REOPOLD  & 50.73 & 26.56 & 15.35 & 1.67 & 3.33 & 0.00 & 16.27 & 74.40 & 57.50 & 31.25 & 3.33 & 13.33 & 0.00 & 29.97 \\
TIP      & 51.68 & 29.69 & 16.45 & 1.25 & 2.08 & 1.25 & 17.07 & 74.60 & \textbf{62.50} & 31.99 & 6.67 & 10.00 & 10.00 & 32.63 \\
\textbf{MetaOPD} & 52.88 & \textbf{30.31} & \textbf{16.68} & \textbf{2.08} & 3.75 & \textbf{2.08} & \textbf{17.96} & \textbf{78.40} & \textbf{62.50} & \textbf{34.56} & \textbf{10.00} & \textbf{16.67} & \textbf{13.33} & \textbf{35.91} \\
\rowcolor{MetaAbsBg}
\textcolor{MetaAbsText}{\bfseries Abs. \(\Delta_{\mathrm{OPD}}\)}
& \textcolor{MetaAbsText}{+2.60} & \textcolor{MetaAbsText}{+5.31} & \textcolor{MetaAbsText}{+0.73} & \textcolor{MetaAbsText}{+1.66} & \textcolor{MetaAbsText}{+0.83} & \textcolor{MetaAbsText}{+0.83} & \textcolor{MetaAbsText}{+1.99}
& \textcolor{MetaAbsText}{+5.40} & \textcolor{MetaAbsText}{+7.50} & \textcolor{MetaAbsText}{+2.94} & \textcolor{MetaAbsText}{+6.67} & \textcolor{MetaAbsText}{+6.67} & \textcolor{MetaAbsText}{+6.66} & \textcolor{MetaAbsText}{+5.97} \\
\rowcolor{MetaRelBg}
\textcolor{MetaRelText}{\bfseries Rel. \(\Delta_{\mathrm{OPD}}\)}
& \textcolor{MetaRelText}{+5.17\%} & \textcolor{MetaRelText}{+21.24\%} & \textcolor{MetaRelText}{+4.58\%} & \textcolor{MetaRelText}{+395.24\%} & \textcolor{MetaRelText}{+28.42\%} & \textcolor{MetaRelText}{+66.40\%} & \textcolor{MetaRelText}{+12.46\%}
& \textcolor{MetaRelText}{+7.40\%} & \textcolor{MetaRelText}{+13.64\%} & \textcolor{MetaRelText}{+9.30\%} & \textcolor{MetaRelText}{+200.30\%} & \textcolor{MetaRelText}{+66.70\%} & \textcolor{MetaRelText}{+99.85\%} & \textcolor{MetaRelText}{+19.94\%} \\
\addlinespace[1.5pt]
\rowcolor{MetaPairBg}
\multicolumn{15}{c}{%
    \textcolor{MetaPairAccent}{\bfseries\textsc{Teacher}}\;\textcolor{MetaPairText}{\bfseries Qwen3-8B}
    \quad\textcolor{MetaPairAccent}{\(\rightsquigarrow\)}\quad
    \textcolor{MetaPairAccent}{\bfseries\textsc{Student}}\;\textcolor{MetaPairText}{\bfseries Qwen3-1.7B-Base}
} \\
\midrule
KD       & 62.93 & 38.13 & 27.02 & 1.25 & 10.00 & 3.75 & 23.85 & 84.80 & 72.50 & 43.38 & 3.33 & 20.00 & 16.67 & 40.11 \\
GRPO     & 66.85 & 37.81 & 28.26 & 1.67 & 8.75 & 5.42 & 24.79 & 83.20 & 72.50 & 48.16 & 3.33 & 23.33 & 16.67 & 41.20 \\
OPD      & 67.23 & 38.75 & 27.80 & 1.67 & 8.33 & 6.67 & 25.07 & 84.20 & 67.50 & 47.43 & 6.67 & 20.00 & 16.67 & 40.41 \\
EOPD     & 67.60 & 40.63 & 27.48 & 1.25 & 10.83 & 7.08 & 25.81 & 83.80 & 72.50 & 44.12 & 6.67 & 30.00 & 20.00 & 42.85 \\
FiRe-OPD & 68.55 & 40.31 & 28.45 & \textbf{2.08} & 7.92 & 7.08 & 25.73 & 85.80 & 72.50 & 43.38 & 6.67 & 16.67 & 23.33 & 41.39 \\
REOPOLD  & 68.85 & 36.25 & \textbf{28.77} & 1.67 & 10.83 & 6.67 & 25.51 & 86.40 & 65.00 & 44.85 & 6.67 & 23.33 & 20.00 & 41.04 \\
TIP      & 68.75 & 41.25 & 27.53 & 1.67 & 11.25 & 6.67 & 26.19 & 86.80 & 72.50 & 45.59 & \textbf{10.00} & 30.00 & \textbf{26.67} & 45.26 \\
\textbf{MetaOPD} & \textbf{69.95} & \textbf{43.75} & 27.71 & \textbf{2.08} & \textbf{12.08} & \textbf{8.33} & \textbf{27.32} & \textbf{88.20} & \textbf{77.50} & \textbf{48.53} & \textbf{10.00} & \textbf{33.33} & 23.33 & \textbf{46.82} \\
\rowcolor{MetaAbsBg}
\textcolor{MetaAbsText}{\bfseries Abs. \(\Delta_{\mathrm{OPD}}\)}
& \textcolor{MetaAbsText}{+2.72} & \textcolor{MetaAbsText}{+5.00} & \textcolor{MetaAbsText}{-0.09} & \textcolor{MetaAbsText}{+0.41} & \textcolor{MetaAbsText}{+3.75} & \textcolor{MetaAbsText}{+1.66} & \textcolor{MetaAbsText}{+2.25}
& \textcolor{MetaAbsText}{+4.00} & \textcolor{MetaAbsText}{+10.00} & \textcolor{MetaAbsText}{+1.10} & \textcolor{MetaAbsText}{+3.33} & \textcolor{MetaAbsText}{+13.33} & \textcolor{MetaAbsText}{+6.66} & \textcolor{MetaAbsText}{+6.41} \\
\rowcolor{MetaRelBg}
\textcolor{MetaRelText}{\bfseries Rel. \(\Delta_{\mathrm{OPD}}\)}
& \textcolor{MetaRelText}{+4.05\%} & \textcolor{MetaRelText}{+12.90\%} & \textcolor{MetaRelText}{-0.32\%} & \textcolor{MetaRelText}{+24.55\%} & \textcolor{MetaRelText}{+45.02\%} & \textcolor{MetaRelText}{+24.89\%} & \textcolor{MetaRelText}{+8.97\%}
& \textcolor{MetaRelText}{+4.75\%} & \textcolor{MetaRelText}{+14.81\%} & \textcolor{MetaRelText}{+2.32\%} & \textcolor{MetaRelText}{+49.93\%} & \textcolor{MetaRelText}{+66.65\%} & \textcolor{MetaRelText}{+39.95\%} & \textcolor{MetaRelText}{+15.86\%} \\
\bottomrule
\end{tabular}%
}
\end{table}


\section{Experiments}
\label{sec:experiments}


\subsection{Experimental Setup}
\label{sec:experimental_setup}

\textbf{Datasets.} We evaluate on six mathematical reasoning benchmarks, \textit{i.e.}, MATH500~\citep{hendrycks2021measuring,lightman2024let}, AMC23~\citep{an2024qwen2}, Minerva Math~\citep{lewkowycz2022solving}, HMMT 2025~\citep{balunovic2026matharena}, AIME24~\citep{aime24}, and AIME25~\citep{zhang2025american}, and assess out-of-domain on three additional datasets, \textit{i.e.}, GPQA-Diamond~\citep{rein2023gpqa}, MMLU Pro~\citep{wang2024mmlu}, and AlpacaEval 2.0~\citep{dubois2024length}. 

\textbf{Baseline methods.} We compare our MetaOPD with seven baselines spanning different learning strategies: KD~\citep{hinton2015distilling}, GRPO~\citep{shao2024deepseekmath}, OPD~\citep{agarwal2024policy}, EOPD~\citep{jin2026entropy}, 
FiRe-OPD~\citep{li2026filter},
REOPOLD~\citep{ko2026scaling}, and TIP~\citep{xu2026tip}.

\textbf{Implementation details.} We use Qwen3~\citep{yang2025qwen3} models, with Qwen3-8B as the frozen teacher and Qwen3-0.6B-Base and Qwen3-1.7B-Base as students. For all methods, both student models are trained for three epochs on MATH~\citep{hendrycks2021measuring}. 
MetaOPD samples rollout prompts and reference solutions from independently shuffled versions of the training split, allowing overlap between the two streams. 
Evaluation examples are excluded from both student and weighting network updates. All methods are evaluated using the same decoding settings and a Math-Verify-based answer verifier. We generate eight responses per problem and report Avg@8, the average accuracy across these responses, and Pass@8, the proportion of problems with at least one correct response. We aggregate each metric using an unweighted average across benchmarks. Full training and evaluation details are provided in Appendix~\ref{app:reproducibility}.

\subsection{Main Results}
\label{sec:main_results}


\textbf{Mathematical-reasoning performance.}
\Cref{tab:main-results} reports the performance of MetaOPD and seven baselines on six mathematical-reasoning benchmarks, using Qwen3-0.6B-Base and Qwen3-1.7B-Base as students. MetaOPD achieves the highest six-benchmark averages for both Avg@8 and Pass@8 at both student scales, including comparisons with adaptive OPD baselines. With the 1.7B student, MetaOPD improves Avg@8/Pass@8 over standard OPD by 2.25/6.41 percentage points and over TIP, which attains the highest aggregate scores among the adaptive OPD baselines, by 1.13/1.56 points. Similar improvements are observed with the 0.6B student, where MetaOPD exceeds OPD by 1.99/5.97 points in Avg@8/Pass@8. The aggregate gains are also broadly reflected at the benchmark level: relative to OPD, MetaOPD improves Pass@8 on all twelve benchmark–scale pairs and Avg@8 on eleven of twelve. The only Avg@8 exception occurs on Minerva Math with the 1.7B student, where MetaOPD is 0.09 points below OPD. On AIME25 with the 1.7B student, MetaOPD obtains the highest Avg@8 while TIP achieves a higher Pass@8, indicating that the two metrics can favor different methods on individual benchmarks.

\begin{table}[t]
\centering
\footnotesize
\definecolor{MetaOODBg}{HTML}{E7F4F2}
\setlength{\tabcolsep}{2pt}
\renewcommand{\arraystretch}{1.10}
\caption{Out-of-domain results for Qwen3-1.7B-Base after MATH training, with Qwen3-8B as the teacher for distillation methods. Base Model is the student before training. Scores are percentages; higher is better. GPQA-Diamond reports Avg@8/Pass@8, MMLU-Pro reports Pass@1, and AlpacaEval 2.0 reports length-controlled win rate (LC-WR) and win rate (WR). Avg. is the unweighted average of these five metric scores. Bold and underline mark the best and second-best results, respectively; the MetaOPD column is shaded.}
\label{tab:ood-results}
\begin{tabular*}{\textwidth}{@{\extracolsep{\fill}}ll*{8}{c}>{\columncolor{MetaOODBg}}c@{}}
\toprule
Benchmark & Metric & \shortstack{Base\\Model} & KD & GRPO & OPD & EOPD & FiRe-OPD & REOPOLD & TIP & \textbf{MetaOPD} \\
\midrule
\multirow{2}{*}{GPQA-Diamond} & Avg@8 & 15.09 & 21.09 & 19.26 & 26.20 & 25.44 & 25.82 & 26.52 & \underline{27.15} & \textbf{27.27} \\
& Pass@8 & 69.19 & 75.25 & 77.27 & \underline{77.78} & 76.26 & 76.77 & 76.26 & 74.75 & \textbf{78.28} \\
\midrule
MMLU-Pro & Pass@1 & 7.28 & 38.13 & 41.86 & 41.57 & 40.93 & 41.51 & 39.52 & \underline{41.90} & \textbf{42.30} \\
\midrule
\multirow{2}{*}{AlpacaEval 2.0} & LC-WR & 1.21 & 14.72 & 13.12 & 15.96 & 15.19 & 15.14 & \underline{16.06} & \textbf{16.70} & 15.98 \\
& WR & 2.03 & 16.09 & 14.63 & 17.97 & 17.17 & 17.13 & 17.37 & \textbf{19.14} & \underline{18.08} \\
\midrule
All metrics & Avg. & 18.96 & 33.06 & 33.23 & 35.90 & 35.00 & 35.27 & 35.15 & \underline{35.93} & \textbf{36.38} \\
\bottomrule
\end{tabular*}
\end{table}


\textbf{Out-of-Domain Generalization.}
We further evaluate the MATH-trained Qwen3-1.7B-Base on GPQA-Diamond, MMLU Pro, and AlpacaEval 2.0, without additional training. \Cref{tab:ood-results} reports the complete results, with the evaluation protocol provided in Appendix~\ref{app:ood-generalization}. MetaOPD achieves the highest scores on both GPQA-Diamond metrics, reaching 27.27 Avg@8 and 78.28 Pass@8, compared with 26.20 and 77.78 for OPD. On MMLU Pro, MetaOPD attains a Pass@1 of 42.30, exceeding OPD by 0.73 percentage points and yielding the highest score among the compared methods. On AlpacaEval 2.0, the improvement over OPD is more limited: MetaOPD obtains 15.98 LC-WR and 18.08 WR, compared with 15.96 and 17.97 for OPD. Overall, these results indicate that the gains from MetaOPD are not confined to the MATH evaluation domain and can transfer to unseen reasoning tasks.

\subsection{Ablations and Controls}
\label{sec:ablations}

\begin{figure*}[t]
    \centering
    \subfloat[Strength sweep\label{fig:metaopd-gold-ablation-a}]{%
        \includegraphics[width=0.5\textwidth,trim=0pt 17pt 334.44pt 29pt,clip]{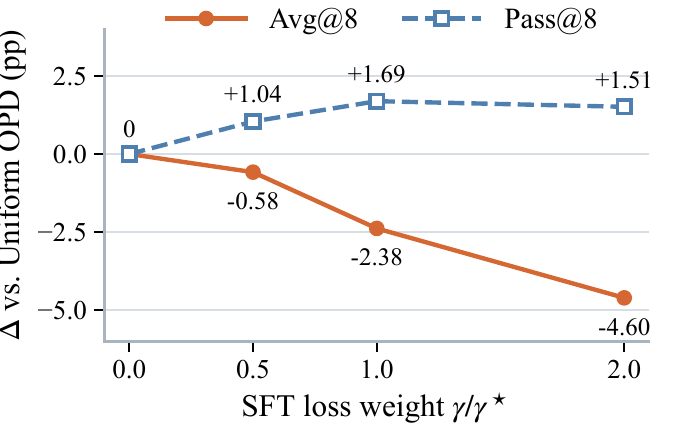}%
    }\hfill
    \subfloat[Effect of adding the SFT loss\label{fig:metaopd-gold-ablation-b}]{%
        \includegraphics[width=0.5\textwidth,trim=334.44pt 17pt 0pt 29pt,clip]{gold_ablation_compact.pdf}%
    }
    \caption{\textbf{Effects of adding SFT to OPD and MetaOPD.}
    Six-benchmark mean Avg@8/Pass@8 changes (percentage points) for Qwen3-1.7B-Base.
    (a) Adding the SFT loss to OPD at half, equal, and double the gradient-matched reference coefficient \(\gamma^\star\approx2.16\).
    (b) Adding the SFT loss at \(\gamma^\star\) to OPD or MetaOPD.
    Changes are relative to each method without the added SFT loss.}
    \label{fig:metaopd-gold-ablation}
\end{figure*}

\textbf{How reference solutions are used.}
We further examine whether the gains of MetaOPD can be attributed to direct supervision from the reference solutions used in the meta-objective. We compare against SFT on reference solutions and variants that add the SFT loss directly to the student objective. The reference coefficient is determined by matching the initial gradient norms of the OPD and SFT losses using training data only, without benchmark-based tuning (Appendix~\ref{app:ablation-resources}). SFT alone underperforms both OPD and MetaOPD in aggregate performance (\cref{tab:metaopd-gold-ablation}). Moreover, adding the SFT loss to uniform OPD increases Pass@8 but consistently reduces Avg@8 across all three tested coefficients (\cref{fig:metaopd-gold-ablation-a}), whereas MetaOPD improves both metrics. Adding the same SFT loss to MetaOPD further decreases Avg@8 and Pass@8 by 5.87 and 6.72 points, respectively (\cref{fig:metaopd-gold-ablation-b}). These controls show that the tested forms of direct supervision do not reproduce MetaOPD's gains in both metrics, supporting the use of validation loss after the student update to optimize the token-weighting rule.


\begin{wraptable}[9]{r}{0.50\textwidth}
\vspace{\dimexpr-\intextsep-1.4pt\relax}
\centering
\footnotesize
\definecolor{MetaAblationBaseText}{HTML}{7A8088}
\definecolor{MetaAblationFullText}{HTML}{168A83}
\definecolor{MetaAblationFullBg}{HTML}{E7F4F2}
\setlength{\tabcolsep}{4pt}
\renewcommand{\arraystretch}{0.92}
\captionsetup{position=top,font=normalsize}
\caption{Feature-ablation Avg. scores (\%).}
\label{tab:metaopd-feature-summary}
\begin{tabularx}{\linewidth}{Xrr}
\toprule
Variant & Avg@8 & Pass@8 \\
\midrule
\textcolor{MetaAblationBaseText}{Uniform OPD} & 25.07 & 40.41 \\
Position only & 25.21 & 39.32 \\
Teacher entropy only & \underline{26.75} & 43.25 \\
Log-probability gap only & 25.42 & \underline{43.62} \\
Top-32 profiles only & 25.70 & 42.24 \\
\rowcolor{MetaAblationFullBg}
\textcolor{MetaAblationFullText}{\textbf{MetaOPD (all features)}} & \textbf{27.32} & \textbf{46.82} \\
\bottomrule
\end{tabularx}
\end{wraptable}

\textbf{Combining token descriptors.}
We examine the contribution of different token descriptors by restricting the weighting network to individual feature groups while keeping its architecture and the meta-training protocol unchanged. The full descriptor achieves the highest aggregate performance, outperforming the strongest single-group variants by 0.57 points in Avg@8 and 3.20 points in Pass@8 \cref{tab:metaopd-feature-summary}. Among the restricted variants, teacher entropy yields the highest Avg@8, whereas the teacher–student log-probability gap yields the highest Pass@8. These results indicate that no individual descriptor fully accounts for the performance of MetaOPD and that combining complementary token-level signals provides a more effective basis for learning the weighting rule. Per-benchmark results and the corresponding input masks are provided in \cref{tab:metaopd-feature-ablation} and Appendix~\ref{app:ablation-resources}.

\section{Conclusion}

On-policy distillation provides dense teacher feedback on student rollouts, but local uncertainty or disagreement alone does not determine whether assigning a token a higher weight within the OPD loss will improve student learning. 
We presented MetaOPD, a one-step bilevel framework that learns to allocate token-level teacher supervision using the student's validation loss after a virtual distillation update. By differentiating this loss on reference solutions through the virtual update, MetaOPD trains a shared weighting network whose token weights guide the student's actual OPD update on its own rollouts.
Across six mathematical-reasoning benchmarks and two Qwen3 student scales, MetaOPD achieves the highest six-benchmark averages for Avg@8 and Pass@8 among the compared methods. Ablations and one-step update comparisons support learning token weights within the OPD loss using feedback on the updated student. Extending these findings to broader models and tasks, relating local update quality to long-horizon learning, and reducing meta-update cost remain open directions.


\subsection*{AI use statement}
OpenAI Codex assisted with research methodology and experimental design, implementation and mathematical checks, manuscript revision and translation, and analysis and presentation of author-supplied results. It did not generate experimental observations or training data. The authors are responsible for verifying the experiments, reviewing all assisted work, and approving the manuscript and artifacts.

\subsection*{Ethics statement}
This work uses existing models and datasets to study on-policy distillation. Distilled students may inherit biases, errors, or harmful behaviors from their teachers; improved benchmark performance does not establish safety. Downstream use should include appropriate safety evaluation and comply with the licenses and terms of use of the underlying models and datasets.

\section*{Reproducibility Statement}

Section~\ref{sec:experiments} specifies the setup and metrics; Appendix~\ref{app:reproducibility}
details configurations and costs, and Appendix~\ref{app:bilevel-math} derives the meta-gradient.
Appendices~\ref{app:intro-motivation-diagnostic}--\ref{app:checkpoint-credit-diagnostic}
give diagnostic protocols, aggregation, and confidence intervals;
Appendix~\ref{app:ood-generalization} reports out-of-domain evaluation.
Code and configurations will accompany the release.

\bibliography{references}
\bibliographystyle{iclr2027_conference}

\newpage
\appendix

\section{Mathematical Details}
\label{app:mathematical-details}

\subsection{Sampled-OPD Gradient and Clipping}
\label{app:sampled-opd-math}

We first expand the log-probabilities used in \cref{eq:sampled-opd-state}. Let \(\mu\) map student action identities to the teacher view, and let \(\mathcal T(s)\) convert a visited student prefix to its teacher context. The conversion preserves prompt tokens and follows the completion-side EOS convention in Appendix~\ref{app:eos-mapping}; stored student actions remain unchanged. Then
\begin{equation}
\begin{aligned}
    u^{\mathrm{old}}_{i,t}
    &=\log\pi_{\theta_{\mathrm{old}}}(a_{i,t}\mid s_{i,t}),\\
    u^\theta_{i,t}
    &=\log\pi_\theta(a_{i,t}\mid s_{i,t}),\\
    u^{\mathrm T}_{i,t}
    &=\log\pi_{\mathrm T}\!\left(\mu(a_{i,t})\mid\mathcal T(s_{i,t})\right).
\end{aligned}
\label{eq:aligned-action-log-probs}
\end{equation}
Each log-probability is the selected logit minus the \(\operatorname{logsumexp}\) over the model's complete vocabulary.

Now fix a visited prefix \(s\), and let \(\mathcal V_{\mathrm S}\) be the student action vocabulary. Because \(\mu\) need not be bijective, the pulled-back teacher scores do not necessarily sum to one over \(\mathcal V_{\mathrm S}\). Define their normalizer and the induced teacher-score distribution as
\[
    Z_{\mu}(s)
    =\sum_{a\in\mathcal V_{\mathrm S}}
      \pi_{\mathrm T}(\mu(a)\mid\mathcal T(s)),
    \qquad
    \pi_{\mathrm T}^{\mu}(a\mid s)
    =\frac{\pi_{\mathrm T}(\mu(a)\mid\mathcal T(s))}{Z_{\mu}(s)}.
\]
Abbreviate \(u^\theta(a)=\log\pi_\theta(a\mid s)\) and
\(u^{\mathrm T}(a)=\log\pi_{\mathrm T}(\mu(a)\mid\mathcal T(s))\). Both \(\mathcal T(s)\) and
\(Z_\mu(s)\) are fixed with respect to the student parameters in this differentiation. Then
\(\log \pi_{\mathrm T}^{\mu}(a\mid s)=u^{\mathrm T}(a)-\log Z_\mu(s)\). The corresponding
full-vocabulary reverse KL is
\[
    \KL\!\left(\pi_\theta(\cdot\mid s)\,\middle\|\,\pi_{\mathrm T}^{\mu}(\cdot\mid s)\right)
    =
    \E_{a\sim\pi_\theta}
    \left[u^\theta(a)-u^{\mathrm T}(a)+\log Z_\mu(s)\right].
\]
Differentiating both the sampling distribution and its log-density gives
\begin{align}
    \nabla_\theta\KL(\pi_\theta\|\pi_{\mathrm T}^{\mu})
    &=
    \E_{a\sim\pi_\theta}
    \left[
        \left(u^\theta(a)-u^{\mathrm T}(a)+\log Z_\mu(s)+1\right)
        \nabla_\theta\log\pi_\theta(a\mid s)
    \right] \nonumber\\
    &=
    \E_{a\sim\pi_\theta}
    \left[
        \left(u^\theta(a)-u^{\mathrm T}(a)\right)
        \nabla_\theta\log\pi_\theta(a\mid s)
    \right],
    \label{eq:appendix-rkl-gradient}
\end{align}
because both \(1\) and \(\log Z_\mu(s)\) are action-independent at the fixed prefix and the expected
score \(\E_{\pi_\theta}[\nabla_\theta\log\pi_\theta(a\mid s)]\) is zero.
At \(\theta=\theta_{\mathrm{old}}\), \(r=1\) and
\[
    \left.\nabla_\theta[-A(a)r(a;\theta)]\right|_{\theta=\theta_{\mathrm{old}}}
    =
    \left(u^{\mathrm{old}}(a)-u^{\mathrm T}(a)\right)
    \left.\nabla_\theta\log\pi_\theta(a\mid s)\right|_{\theta=\theta_{\mathrm{old}}},
\]
so a behavior-policy action provides an unbiased single-sample estimator of
\cref{eq:appendix-rkl-gradient}. This statement is conditional on the visited prefix and does not
differentiate the trajectory distribution.

Away from the rollout policy, the objective with the cached \(A\) held fixed has expectation
\[
    \E_{a\sim\pi_{\theta_{\mathrm{old}}}}\!\left[-A(a)r(a;\theta)\right]
    =
    \E_{a\sim\pi_\theta}\!\left[u^{\mathrm{old}}(a)-u^{\mathrm T}(a)\right].
\]
It therefore differs from the current reverse KL, whose score coefficient contains
\(u^\theta(a)-u^{\mathrm T}(a)\). The importance ratio supports a local update from cached
behavior-policy actions; it does not restore exact reverse-KL equivalence at an arbitrary
\(\theta\).

The active branch of the clipped loss depends on the sign of \(A\). For \(A>0\), it is
\(-A\min(r,1+\epsilon_h)\), so an excessive probability increase is clipped. For \(A<0\), it is
\(-A\max(r,1-\epsilon_l)\), so an excessive probability decrease is clipped. At a clipping
boundary, automatic differentiation supplies the implementation-defined subgradient.

\subsection{Score Centering and Response Normalization}
\label{app:weight-normalization-math}

For response \(i\), let \(n_i=\sum_t m_{i,t}\). Under the convention following \cref{eq:weight-normalization}, all normalized weights are zero when \(n_i=0\). Below, assume \(n_i>0\) and define
\[
    \overline w_i=\frac{1}{n_i}\sum_t m_{i,t}\widetilde w_{i,t}.
\]
Then \cref{eq:weight-normalization} is \(w_{i,t}=\widetilde w_{i,t}/\overline w_i\) on valid
positions, which immediately gives
\[
    \frac{1}{n_i}\sum_t m_{i,t}w_{i,t}=1.
\]
With \(\widetilde w\in(1-\rho,1+\rho)\), the normalized weights satisfy
\[
    w_{i,t}\in
    \left(\frac{1-\rho}{1+\rho},\frac{1+\rho}{1-\rho}\right).
\]
With \(\rho=0.25\), the normalized range is \(w_{i,t}\in(0.6,5/3)\), even though the final
weights need not remain inside the raw interval \((0.75,1.25)\).

For valid positions \(t\) and \(j\),
\[
    \frac{\partial w_{i,t}}{\partial\widetilde w_{i,j}}
    =
    \frac{\mathbb I[t=j]}{\overline w_i}
    -
    \frac{\widetilde w_{i,t}}{n_i\overline w_i^2},
\]
which makes the within-response coupling explicit. Combining this derivative with
\cref{eq:weight-hypergradient} gives the gradient with respect to the unnormalized weights in
\cref{eq:relative-token-credit}, expanded here for a valid raw weight:
\[
    \frac{\partial\mathcal L_{\mathrm{out}}}{\partial\widetilde w_{i,j}}
    =-
    \frac{1}{N_B\overline w_i}
    \left(
        c_{i,j}-\frac{1}{n_i}\sum_t m_{i,t}w_{i,t}c_{i,t}
    \right).
\]
Here \(c_{i,t}\) is defined in \cref{eq:weight-hypergradient}; its
learning-rate dependence is already included in \(\boldsymbol d\). Thus normalization subtracts
the response-level weighted mean from \(c_{i,j}\), rather than using its sign alone.
Invalid weights are zero, so a completely masked response contributes to
neither the weighted loss nor the weight regularizer. The implementation also clamps valid counts
below by one and rejects optimizer minibatches with \(N_B=0\).

Score centering introduces a second coupling before the residual map. On the valid positions
of response \(i\), let \(\boldsymbol z_i\in\mathbb R^{n_i}\) collect the network scores and define
\[
 P_i=I-\frac{1}{n_i}\boldsymbol 1\boldsymbol 1^\top,
 \qquad \boldsymbol z_i^\circ=P_i\boldsymbol z_i,
 \qquad D_i=\operatorname{diag}\!\left(\rho[1-\tanh^2(z_{i,t}^\circ)]\right).
\]
Writing \(J_i\) for the normalization Jacobian above, the complete derivative is
\begin{equation}
 \frac{\partial\boldsymbol w_i}{\partial\boldsymbol z_i}=J_iD_iP_i,
 \qquad
 \nabla_{\boldsymbol z_i}\mathcal L_{\mathrm{out}}
 =-\frac{1}{N_B}P_iD_iJ_i^\top\boldsymbol c_i.
 \label{eq:centered-score-jacobian}
\end{equation}
Both \(P_i\) and \(D_i\) are symmetric. In particular, the score derivative includes the
mean-score derivative; detaching that mean would change the meta-gradient. A common score
shift is annihilated by \(P_i\), and a response with one valid token has constant weight one.
The same Jacobian propagates the regularizer derivative, with
\(\nabla_{\boldsymbol w_i}\mathcal R_w=2(\boldsymbol w_i-\boldsymbol 1)/N_B\).
The implementation safeguards the normalization denominator by \(10^{-8}\); since
\(\sum_t m_{i,t}\widetilde w_{i,t}>n_i(1-\rho)\), this safeguard is inactive for nonempty responses.

\subsection{AdamW Virtual Update}
\label{app:virtual-adamw}

Consider one optimizer minibatch and suppress its training index. For each trainable parameter
tensor \(\theta_j\), the current real optimizer supplies first and second moments
\(\boldsymbol m_j,\boldsymbol v_j\), a completed-step count \(\tau_j\), and its parameter-group
settings \((\alpha_j,\beta_{1,j},\beta_{2,j},\varepsilon_j,\lambda_j)\): learning rate, moment
decays, denominator epsilon, and decoupled weight decay. These quantities form \(\mathsf S\) in
\cref{eq:virtual-update}. The virtual step reads them as constants with respect to \(\eta\).
Zero initialization is used only if the corresponding real optimizer state is uninitialized;
moments and step counts are not reset at subsequent virtual steps. We assume the standard
settings \(0\leq\beta_{1,j}<1\), \(0<\beta_{2,j}<1\), and \(\varepsilon_j>0\).

\textbf{Global gradient clipping.}
The input to clipping is the full-model gradient of the global token-mean inner loss,
after microbatch accumulation and distributed reduction. Let \(G_{\max}>0\) be the real update's
norm threshold and \(\varepsilon_{\mathrm{clip}}\geq0\) its numerical safeguard. For
\(n=\|\boldsymbol g_{\mathrm{in}}(\eta)\|_2>0\), define
\begin{equation}
    \kappa_{\mathrm{clip}}=\min\!\left(1,\frac{G_{\max}}{n+\varepsilon_{\mathrm{clip}}}\right),
    \qquad
    \overline{\boldsymbol g}_{\mathrm{in}}(\eta)=\mathcal C(\boldsymbol g_{\mathrm{in}}(\eta))
    =\kappa_{\mathrm{clip}}\boldsymbol g_{\mathrm{in}}(\eta).
\label{eq:appendix-global-clip}
\end{equation}
The zero gradient maps to zero. Clipping uses one norm over all trainable student parameters,
including parameters in different optimizer groups; it is not applied separately to each
microbatch or parameter tensor. On a smooth branch, its Jacobian is
\begin{equation}
    J_{\mathcal C}=
    \begin{cases}
        I, & n+\varepsilon_{\mathrm{clip}}<G_{\max},\\[2pt]
        \displaystyle
        \kappa_{\mathrm{clip}} I-\frac{G_{\max}}{n(n+\varepsilon_{\mathrm{clip}})^2}
        \boldsymbol g_{\mathrm{in}}(\eta)\boldsymbol g_{\mathrm{in}}(\eta)^\top,
        & n+\varepsilon_{\mathrm{clip}}>G_{\max},\ n>0.
    \end{cases}
\label{eq:appendix-clip-jacobian}
\end{equation}
At the switching boundary, the derivative follows the branch convention of the real clipping
operation. The rank-one term accounts for the dependence of the clipping coefficient on the
weights through \(\boldsymbol g_{\mathrm{in}}(\eta)\). Detaching that coefficient would omit this term.
At the origin, \(J_{\mathcal C}=\min(1,G_{\max}/\varepsilon_{\mathrm{clip}})I\) for
\(\varepsilon_{\mathrm{clip}}>0\), and \(J_{\mathcal C}=I\) when the safeguard is zero.
If clipping is disabled, \(\mathcal C\) and \(J_{\mathcal C}\) are the identity.

\textbf{Moment update and parameter map.}
For standard AdamW, the virtual moments, bias corrections, and virtual student parameters are
\begin{equation}
\begin{aligned}
    \boldsymbol m_j^+
    &=\beta_{1,j}\boldsymbol m_j+(1-\beta_{1,j})\overline{\boldsymbol g}_{\mathrm{in},j}(\eta),\\
    \boldsymbol v_j^+
    &=\beta_{2,j}\boldsymbol v_j+(1-\beta_{2,j})\overline{\boldsymbol g}_{\mathrm{in},j}(\eta)^{\,2},\\
    \widehat{\boldsymbol m}_j
    &=\frac{\boldsymbol m_j^+}{1-\beta_{1,j}^{\,\tau_j+1}},
    &\widehat{\boldsymbol v}_j
    &=\frac{\boldsymbol v_j^+}{1-\beta_{2,j}^{\,\tau_j+1}},\\
    \theta'_j
    &=(1-\alpha_j\lambda_j)\theta_j
      -\alpha_j\frac{\widehat{\boldsymbol m}_j}
      {\sqrt{\widehat{\boldsymbol v}_j}+\varepsilon_j}.
\end{aligned}
\label{eq:appendix-adamw-map}
\end{equation}
All products, squares, square roots, and divisions within a parameter tensor are elementwise.
The learning rate \(\alpha_j\) is read from the current real parameter group, including scheduled
decay. Bias correction uses that tensor's next step \(\tau_j+1\). The expressions above create
temporary values; they do not overwrite \(\theta_j\), \(\boldsymbol m_j\), \(\boldsymbol v_j\),
or \(\tau_j\). Only the later real student step advances the stored moments and step counts.

For clarity, the AdamW Jacobian can also be written coordinatewise. Let
\[
    \zeta_{1,j}=\frac{1-\beta_{1,j}}{1-\beta_{1,j}^{\,\tau_j+1}},\qquad
    \zeta_{2,j}=\frac{1-\beta_{2,j}}{1-\beta_{2,j}^{\,\tau_j+1}},\qquad
    \sigma_{j,k}=\sqrt{\widehat v_{j,k}},
\]
where \(k\) indexes a coordinate of tensor \(j\). For \(\sigma_{j,k}>0\),
\begin{equation}
    \frac{\partial\theta'_{j,k}}{\partial\overline g_{\mathrm{in},j,k}(\eta)}
    =-\alpha_j\left[
        \frac{\zeta_{1,j}}{\sigma_{j,k}+\varepsilon_j}
        -\frac{\zeta_{2,j}\widehat m_{j,k}\overline g_{\mathrm{in},j,k}(\eta)}
        {\sigma_{j,k}(\sigma_{j,k}+\varepsilon_j)^2}
    \right].
\label{eq:appendix-adamw-jacobian}
\end{equation}
The off-diagonal entries of \(J_{\mathcal A}\) are zero. The second term differentiates the
new second moment; treating AdamW as a fixed preconditioner would discard it. At a zero-gradient
coordinate with zero first and second moments, the composite map has derivative
\(-\alpha_j\zeta_{1,j}/\varepsilon_j\), obtained by continuity rather than division by
\(\sigma_{j,k}=0\). Decoupled weight decay has no direct derivative with respect to the gradient
input because \(\theta_j\) is fixed. It still changes \(\theta'\), and hence the outer gradient,
so it must remain in the forward virtual map.

\subsection{One-Step Hypergradient and VJP--JVP Form}
\label{app:bilevel-math}

We derive the token decomposition in \S\ref{sec:token-credit} from the complete meta-gradient in \cref{eq:bilevel-meta-gradient}. Vectorize the student and weighting parameters as \(\theta\in\R^p\) and \(\eta\in\R^q\). Gradients are column vectors; the Jacobian of a map from \(\R^q\) to \(\R^p\) has shape \(p\times q\), with rows indexing outputs and columns indexing inputs. We use the fixed-input convention of \S\ref{sec:bilevel-reweighting}: current student parameters, real optimizer state, rollout actions, masks, old/teacher caches, and detached descriptors are held fixed.

Only the normalized weights in \cref{eq:inner-gradient-dependence} depend on \(\eta\). Since \(\boldsymbol g_{i,t}\in\R^p\) and \(\nabla_\eta w_{\eta,i,t}\in\R^q\), differentiating each weighted term gives the outer product
\begin{equation}
    G_\eta:=\frac{\partial\boldsymbol g_{\mathrm{in}}(\eta)}{\partial\eta}
    =\frac{1}{N_B}\sum_{i,t}m_{i,t}
      \boldsymbol g_{i,t}\left(\nabla_\eta w_{\eta,i,t}\right)^\top
    \in\R^{p\times q},
    \qquad
    \frac{\partial\theta'}{\partial\eta}=J_{\mathcal A}J_{\mathcal C}G_\eta.
\label{eq:appendix-inner-jacobian}
\end{equation}
The two optimizer Jacobians are \(p\times p\), and \(\boldsymbol g_{\mathrm{out}}\in\R^p\). Thus the validation-loss term of \cref{eq:bilevel-meta-gradient} is the \(q\)-vector
\begin{equation}
    \nabla_\eta\mathcal L_{\mathrm{out}}
    =\left(\frac{\partial\boldsymbol g_{\mathrm{in}}(\eta)}{\partial\eta}\right)^\top
       J_{\mathcal C}^{\top}J_{\mathcal A}^{\top}\boldsymbol g_{\mathrm{out}}
    =G_\eta^\top J_{\mathcal C}^{\top}J_{\mathcal A}^{\top}\boldsymbol g_{\mathrm{out}}.
\label{eq:meta-chain-rule}
\end{equation}
Using \(J_{\mathcal C}^{\top}J_{\mathcal A}^{\top}\boldsymbol g_{\mathrm{out}}=-\boldsymbol d\) from \cref{eq:optimizer-aware-direction} and expanding \(G_\eta^\top\) yields
\begin{equation}
    \nabla_\eta\mathcal L_{\mathrm{out}}
    =-\frac{1}{N_B}\sum_{i,t}m_{i,t}
      \left(\nabla_\eta w_{\eta,i,t}\right)\boldsymbol g_{i,t}^{\top}\boldsymbol d
    =-\frac{1}{N_B}\sum_{i,t}m_{i,t}c_{i,t}\nabla_\eta w_{\eta,i,t}.
\label{eq:appendix-adamw-hypergradient}
\end{equation}
Here \(\boldsymbol d\in\R^p\) and \(c_{i,t}=\boldsymbol g_{i,t}^{\top}\boldsymbol d\) is scalar, establishing \cref{eq:weight-hypergradient}.

To verify the gradient of \cref{eq:compatibility-surrogate}, fix an evaluation point \(\eta_0\) and write \(c^0_{i,t}=c_{i,t}(\eta_0)\). The VJP--JVP procedure in \S\ref{sec:token-credit} supplies these values. Differentiating the objective while holding them fixed gives
\[
\begin{aligned}
    \left.\nabla_\eta\widehat{\mathcal L}_{\mathrm{meta}}^{(r)}(\eta;B,\boldsymbol c^0)\right|_{\eta_0}
    &=-\frac{1}{N_B}\sum_{i,t}m_{i,t}c^0_{i,t}
      \left.\nabla_\eta w_{\eta,i,t}\right|_{\eta_0}
      +\lambda_w^{(r)}\left.\nabla_\eta\mathcal R_w\right|_{\eta_0}\\
    &=\left.\nabla_\eta\mathcal J_{\mathrm{meta}}\right|_{\eta_0}.
\end{aligned}
\]
The second equality uses \cref{eq:appendix-adamw-hypergradient}. Exactness concerns the specified single-step map, including its new moments and clipping coefficient, under the fixed-input convention above. At clipping boundaries, the derivative follows the corresponding automatic-differentiation convention.

\section{Implementation and Reproducibility Details}
\label{app:reproducibility}

The inner rollout stream and the reference-solution stream are separately shuffled iterators over the MATH training split.
They are not constrained to be sample-disjoint, and the experimental setup does not create a separate validation pool for meta-updates. Evaluation examples do not update either the student parameters \(\theta\) or the weighting network parameters \(\eta\).

\subsection{Token Descriptor}

\Cref{tab:descriptor} expands the 74-dimensional descriptor defined in \cref{sec:token-weighted-opd}, following its detachment and sampled-loss conventions. Top-\(k_f\) token identities are used only to compute the top-1 match and top-\(k_f\) overlap; the identities themselves are not network inputs.

\begin{table}[!htb]
\centering
\small
\setlength{\tabcolsep}{4pt}
\renewcommand{\arraystretch}{1.08}
\caption{Composition of the 74-dimensional weighting network descriptor, with \(k_f=32\).}
\label{tab:descriptor}
\begin{tabular}{@{}p{0.20\textwidth}p{0.58\textwidth}cc@{}}
\toprule
Feature group & Features & Dim. & Detached \\
\midrule
Sampled action
& Current-student log-probability, teacher log-probability, and teacher-minus-current log-probability gap
& 3 & Yes \\
Uncertainty
& Vocabulary-normalized student and teacher entropy
& 2 & Yes \\
Top-1 behavior
& Student and teacher top-1 probability; student--teacher top-1 identity match
& 3 & Yes \\
Position and overlap
& Position normalized by the common completion-window width; student--teacher top-32 set overlap
& 2 & Yes \\
Ranked probabilities
& Student top-32 and teacher top-32 full-vocabulary-normalized log-probability values
& 64 & Yes \\
\midrule
Total & Ten scalar statistics plus two ranked top-32 vectors & 74 & Yes \\
\bottomrule
\end{tabular}
\end{table}

For a common completion tensor of width \(L_{\mathrm{win}}\), the position feature is zero when \(L_{\mathrm{win}}=1\) and otherwise equals \((t-1)/(L_{\mathrm{win}}-1)\), using the one-based token index in \cref{eq:old-policy-rollout}. The same position grid is shared across rows before masking; it is not renormalized by each response's number of valid completion tokens.

Let \(p^{\mathrm S}\) and \(p^{\mathrm T}\) be the full-vocabulary student and teacher distributions
at their respective context views. The entropy features are
\[
    h^{\mathrm S}=\frac{-\sum_{a\in\mathcal V_{\mathrm S}}p^{\mathrm S}(a)\log p^{\mathrm S}(a)}
                          {\log|\mathcal V_{\mathrm S}|},
    \qquad
    h^{\mathrm T}=\frac{-\sum_{a\in\mathcal V_{\mathrm T}}p^{\mathrm T}(a)\log p^{\mathrm T}(a)}
                          {\log|\mathcal V_{\mathrm T}|}.
\]
Let \(S_{32}\) and \(T_{32}\) contain the top-32 token identities in the two distributions.
With teacher identities canonicalized as in Appendix~\ref{app:eos-mapping}, the overlap feature is
\[
    \operatorname{Overlap}_{32}
    =\frac{1}{32}\sum_{a\in S_{32}}
       \mathbb I\!\left[a\in\operatorname{Canon}(T_{32})\right].
\]
The denominator remains 32 even if canonicalization merges teacher identities; this is a
student-set coverage fraction, not Jaccard overlap.

The weighting network has linear widths 74--256--64--1, SiLU activations after the first two linear layers, and dropout probability zero. Its final scalar head is zero-initialized. Consequently, the initial raw and response-normalized weights are exactly one on all valid tokens. The raw weights use \(\rho=0.25\), and the normalized-weight regularizer starts at \(\lambda_w^{(0)}=10^{-3}\) and follows Appendix~\ref{app:meta-schedule}.

\subsection{Training and Evaluation Configuration}

\Cref{tab:metaopd-reproducibility} summarizes the shared training and evaluation configuration for the main and ablation experiments.
Settings for the weighting network and virtual updates apply only to variants with meta-updates. Settings for reference-solution data apply to variants that use reference solutions for supervision.
All learned-weight variants use response-centered scores and the schedules below, except for the factor explicitly changed in an ablation.
The ablations use the same optimizer settings as the main experiments; virtual student steps follow the actual student's AdamW update rule.

\textbf{Baseline settings.}
KD distills the student on teacher-generated sequences. GRPO uses standard answer-correctness rewards with a group size of eight. EOPD uses the method-specific settings reported in its paper. Appendix~\ref{app:token-allocation-baselines} details how the FiRe-OPD, REOPOLD, and TIP allocation rules are implemented within the shared sampled-OPD framework.

\begin{table}[t]
\centering
\small
\setlength{\tabcolsep}{6pt}
\renewcommand{\arraystretch}{1.08}
\caption{Shared training and evaluation configuration, including MetaOPD-specific settings.}
\label{tab:metaopd-reproducibility}
\begin{tabular}{@{}p{0.34\textwidth}p{0.60\textwidth}@{}}
\toprule
Setting & Value \\
\midrule
Student / teacher models & Qwen3-0.6B-Base or Qwen3-1.7B-Base / frozen Qwen3-8B \\
Training data / MetaOPD reference-solution data & MATH train / MATH train \\
MetaOPD rollout / reference sampling & Separate shuffled streams; overlap allowed; no meta split \\
Shared epochs / training seed & 3 / 42 \\
Shared rollout and update schedule & 174 rollout iterations; four student updates per rollout; 696 student updates \\
Shared batch sizes & 128 rollout / 32 student optimizer; global reference batch 32 where used \\
Shared student optimizer / learning rate & AdamW / \(3\times10^{-6}\), cosine decay \\
MetaOPD meta optimizer / initial learning rate & AdamW / \(2.5\times10^{-5}\); scaled by the student learning-rate ratio \\
MetaOPD virtual optimizer / state & Same AdamW rule as the actual student update; current student optimizer state, without committing changes \\
MetaOPD virtual learning rates & Current real student optimizer rates, including scheduled decay \\
Student gradient clipping & Full-model global norm, threshold 1.0 in virtual and actual steps \\
Sampled-OPD ratio clipping & \((\epsilon_l,\epsilon_h)=(0.2,0.2)\) \\
MetaOPD residual / initial regularizer & \(\rho=0.25\) / \(\lambda_w^{(0)}=10^{-3}\); regularizer follows the student learning-rate ratio \\
MetaOPD score centering / meta updates & Valid-response mean subtracted before \(\tanh\) / one per student update \\
Shared training decoding & Temperature 1.0; top-\(p\) 1.0; maximum completion 4096 \\
Math evaluation decoding & Eight samples; temperature 1.0; top-\(p\) 0.8; maximum response 8192 \\
Allocation for reported GPU-hours & Eight NVIDIA A100 80GB GPUs: four student DDP and four paired teacher GPUs \\
\bottomrule
\end{tabular}
\end{table}

\subsubsection{Token-Allocation Baseline Implementations}
\label{app:token-allocation-baselines}

To study supervision allocation under a common OPD training framework, we implement the token-selection and weighting rules of FiRe-OPD~\citep{li2026filter}, REOPOLD~\citep{ko2026scaling}, and TIP~\citep{xu2026tip} on top of the sampled-OPD surrogate in \cref{eq:sampled-opd-token-loss}. The underlying surrogate form, student optimizer, and training schedule are shared; method-specific selection, weighting, reward processing, and loss reduction are specified below. This design focuses the comparison on supervision allocation while controlling for differences in the original methods' underlying OPD objectives.

\textbf{Shared implementation.}
Selection statistics use the frozen rollout-time student and teacher predictions and are detached from the student update. Batch statistics and rankings are computed over valid completion tokens in the full 32-response optimizer minibatch, across all microbatches and student ranks; response-level selection is identified below. The models, data, rollout batch size, ratio clipping, and decoding follow \cref{tab:metaopd-reproducibility}. All three baselines use AdamW with zero weight decay, cosine decay advanced once per rollout, BF16 forward computation, and FP32 student parameters and sampled-action log-probabilities. They use no learned weighting network, meta-update, or reference-solution loss. Training runs for 696 student updates, and evaluation uses the final checkpoint at update 696.

\begin{table}[htbp]
\centering
\small
\setlength{\tabcolsep}{6pt}
\renewcommand{\arraystretch}{1.12}
\caption{Method-specific allocation settings within the shared sampled-OPD framework. Selection scopes and loss normalization are specified in the text.}
\label{tab:token-allocation-baseline-settings}
\begin{tabular}{@{}p{0.18\textwidth}p{0.76\textwidth}@{}}
\toprule
Method & Configuration \\
\midrule
FiRe-OPD & Trajectory drop fraction 0.2; teacher-confidence and student-confusion coefficients both 1.0 \\
REOPOLD & Reward parameter \(\lambda=0.3\); phase switch at update 232; phase-II entropy keep fraction 0.2 \\
TIP & Soft-OR; response-level keep fraction 0.5; entropy clipping at the 98th percentile; exact-KL computation in 64-token chunks \\
\bottomrule
\end{tabular}
\end{table}

\textbf{FiRe-OPD.}
Responses are ranked by mean teacher log-probability of their sampled tokens. We remove the lowest-scoring \(\lfloor0.2|B|\rfloor\) responses (six for \(|B|=32\)). Let \(h^{\mathrm T}\) and \(h^{\mathrm S}\) denote vocabulary-normalized teacher and rollout-student entropies. Before filtering, their respective maxima are computed over all valid tokens in \(B\). Teacher confidence and student confusion are
\[
    c^{\mathrm T}_{i,t}=1-\frac{h^{\mathrm T}_{i,t}}{\max_B h^{\mathrm T}},
    \qquad c^{\mathrm S}_{i,t}=\frac{h^{\mathrm S}_{i,t}}{\max_B h^{\mathrm S}}.
\]
Retained-token weights are proportional to \((1+c^{\mathrm T}_{i,t})(1+c^{\mathrm S}_{i,t})\), normalized to mean one within each retained response. Entropy denominators are lower-bounded by FP32 machine epsilon; removed responses contribute neither loss nor denominator tokens.

\textbf{REOPOLD.}
For the detached reward \(R_{i,t}=u^{\mathrm T}_{i,t}-u^{\mathrm{old}}_{i,t}\), define \(R_{\min}=\log(\lambda)/(1-\lambda)\). We replace \(A_{i,t}\) in \cref{eq:sampled-opd-token-loss} with \(\max(R_{i,t},R_{\min})\). At zero-based updates 0--231, tokens with \(R_{i,t}\ge R_{\min}\) are retained. At updates 232--695, selection instead retains the \(\lceil0.2N_B\rceil\) highest-student-entropy tokens across the optimizer minibatch. If a response would be empty, its highest-reward token in phase I or highest-entropy token in phase II is retained. This safeguard can raise the actual retained fraction above 0.2 in phase II.

\textbf{TIP Soft-OR.}
The two signals are \(h_{i,t}=H(\pi_{\theta_{\mathrm{old}}}(\cdot\mid s_{i,t}))/\log|\Vocab|\) and the exact full-vocabulary \(d_{i,t}=\KL(\pi_{\theta_{\mathrm{old}}}\Vert\pi_{\mathrm T})\) at the corresponding student and teacher context views. The latter is computed from full logits, rather than from the sampled-token log-probability gap. We cap \(h\) at its minibatch 98th percentile, then separately min--max normalize the capped entropy and \(d\) to obtain \(\hat h\) and \(\hat d\). A signal whose range is at most FP32 machine epsilon maps to zero. The score is \(s_{i,t}=\hat h_{i,t}+\hat d_{i,t}-\hat h_{i,t}\hat d_{i,t}\). Each response retains its \(\max(1,\lfloor0.5n_i\rfloor)\) highest-scoring valid tokens. Ranking ties in all three methods preserve the gathered input order. Full-vocabulary reverse KL is used for token selection; the student update on retained tokens uses the shared sampled-OPD surrogate in \cref{eq:sampled-opd-token-loss} in place of TIP's original full-vocabulary reverse-KL training loss.

\textbf{Reduction after selection.}
For all three baselines, the loss is the weighted sum of retained-token surrogate losses divided by the total number of retained tokens in the global optimizer minibatch. FiRe-OPD uses its response-normalized soft weights; REOPOLD and TIP use unit weights on retained tokens and zero weights elsewhere. The binary masks are not rescaled to have mean one over the original response. Thus, filtering changes both the numerator and the token-count denominator.

\subsection{Global Token-Mean Reduction}

Both the virtual and actual student updates evaluate the weighted OPD loss using the global valid-token count \(N_B\) from the complete 32-example optimizer minibatch. The numerator is accumulated across all local microbatches and distributed ranks before division by \(N_B\); the implementation does not average independently normalized microbatch losses. Response-level weight normalization in \cref{eq:weight-normalization} is applied before this global reduction.

To compute validation loss, we similarly sum token losses on reference solutions over the microbatches and participating student ranks, then divide by the global valid reference-solution token count \(N_{\mathrm{ref}}\). Because completion lengths differ, \(N_B\) and \(N_{\mathrm{ref}}\) are generally different even though both global example batches contain 32 examples.
The inner gradient is globally reduced before norm clipping and the virtual optimizer map; the outer gradient is globally reduced before the optimizer VJPs. Clipping each rank's partial gradient would define a different virtual update.

\subsection{Virtual State and Learning-Rate Convention}

As specified in \S\ref{sec:bilevel-reweighting}, training-time virtual steps use the current student AdamW state and scheduled learning rates without committing optimizer or parameter changes.

The motivating comparison in Appendix~\ref{app:intro-motivation-diagnostic} starts from zero student moments and step zero. The checkpoint probes in Appendix~\ref{app:checkpoint-credit-diagnostic} restore the trained student optimizer state. Both use a fixed reference learning rate of \(3\times10^{-6}\), whereas training-time virtual steps read the scheduled rate.

\textbf{Coupled learning-rate schedules.}
\label{app:meta-schedule}
Let \(\alpha_g^{(r)}\) be the live student learning rate in trainable parameter group \(g\)
at global student step \(r\), and let \(\alpha_0=3\times10^{-6}\) be the configured initial rate.
The schedule scale and effective meta settings are
\begin{equation}
    \chi_r=\min\!\left(1,\max\!\left(0,\max_g\frac{\alpha_g^{(r)}}{\alpha_0}\right)\right),
    \quad \alpha_\eta^{(r)}=\alpha_{\eta,0}\chi_r,
    \quad \lambda_w^{(r)}=\lambda_w^{(0)}\chi_r,
 \label{eq:meta-schedule}
\end{equation}
where \(\alpha_{\eta,0}=2.5\times10^{-5}\) and \(\lambda_w^{(0)}=10^{-3}\).
The student uses cosine decay, advanced once per rollout round; all four optimizer minibatches
in that round read its current rate. The weighting network uses AdamW with zero weight decay and is
updated at each student step throughout all 696 updates. The schedule values are constants
for the current one-step meta-gradient. Neither the virtual step nor the meta-update advances
the student scheduler.

\subsection{End-of-Sequence Mapping}
\label{app:eos-mapping}

The action map \(\mu\) in \cref{eq:aligned-action-log-probs} is the identity except for
\(\mu(151643)=151645\), mapping the student EOS identity to the teacher EOS identity.
The context conversion \(\mathcal T\) leaves prompt tokens unchanged and applies this mapping
only to the first valid completion occurrence of 151643 in the teacher view. Padding, later
invalid positions, and stored student actions are unchanged. The valid sampled EOS action is
scored with its teacher identity. Teacher top-\(k\) identities are canonicalized back to the
student identity before match and overlap features are computed. This convention aligns EOS
across model views without introducing a dual-EOS student stop set.

Reference-solution targets use the student EOS token 151643, which is also the sole generation stop token. The teacher-view mapping does not change this student target identity. The SFT, OPD + SFT, and MetaOPD + SFT controls use this same EOS convention.

\subsection{Ablation and Resource Details}
\label{app:ablation-resources}

\textbf{Variant definitions and SFT loss calibration.}
Let \(\mathcal L_{\mathrm{OPD}}^w\) denote the OPD loss on student-generated responses with token weights \(w\). In OPD + SFT, every valid token has weight one. In MetaOPD + SFT, the weighting network assigns the weights, which have mean one over the valid tokens in each response. Let \(\mathcal L_{\mathrm{SFT}}\) denote the supervised fine-tuning (SFT) loss: the mean token prediction loss on reference solutions under teacher forcing. It has the form of \cref{eq:outer-loss}, evaluated at the current student parameters rather than the virtual parameters. Each loss is averaged over the valid tokens in its own global batch. Both variants update the student with
\[
    \mathcal L_{\mathrm{student}}
    =\mathcal L_{\mathrm{OPD}}^w+\gamma\mathcal L_{\mathrm{SFT}}.
\]

The coefficient is calibrated once at the common Qwen3-1.7B-Base initialization using 16 disjoint pairs of 32-example OPD and reference-solution batches:
\[
    \gamma^\star
    =\operatorname{median}_{k=1}^{16}
    \frac{\lVert\nabla_\theta\mathcal L_{\mathrm{OPD}}^{(k)}\rVert_2}
         {\lVert\nabla_\theta\mathcal L_{\mathrm{SFT}}^{(k)}\rVert_2+10^{-12}}.
\]
Thus \(\gamma^\star\) approximately matches the two gradient scales at initialization; it is applied only to the SFT loss in the student objective, not to the MetaOPD outer objective.

The resulting coefficient is \(\gamma^\star\approx2.16\); the sweep uses half, equal, and double this value. Calibration uses training data only and is not selected by benchmark performance. Matching gradient norms does not imply that the two gradients have the same direction or induce equal optimizer updates.

For MetaOPD + SFT, the virtual inner objective remains weighted sampled OPD. The same global reference batch used for the outer objective supplies the SFT loss in the subsequent real student update; this loss is evaluated at the real student parameters.

\begin{table}[t]
\centering
\small
\setlength{\tabcolsep}{5pt}
\renewcommand{\arraystretch}{1.08}
\caption{Ablation-variant mapping. ``Outer'' means that reference solutions update only the weighting network through the meta objective; ``direct'' means that their supervised loss enters the student objective.}
\label{tab:ablation-variants}
\begin{tabular}{@{}p{0.18\textwidth}p{0.24\textwidth}p{0.13\textwidth}p{0.34\textwidth}@{}}
\toprule
Variant & Token weighting & Reference use & Student objective \\
\midrule
Uniform OPD & Uniform & None & \(\mathcal L_{\mathrm{OPD}}\) \\
Position only & Position & Outer & \(\mathcal L_{\mathrm{OPD}}^w\) \\
Teacher entropy & Teacher entropy & Outer & \(\mathcal L_{\mathrm{OPD}}^w\) \\
Log-probability gap & \(\delta_{i,t}\) & Outer & \(\mathcal L_{\mathrm{OPD}}^w\) \\
Top-32 profiles & Student + teacher profiles & Outer & \(\mathcal L_{\mathrm{OPD}}^w\) \\
MetaOPD & All 74 descriptors & Outer & \(\mathcal L_{\mathrm{OPD}}^w\) \\
SFT & None & Direct only & \(\mathcal L_{\mathrm{SFT}}\) \\
OPD + SFT (half) & Uniform & Direct only & \(\mathcal L_{\mathrm{OPD}}+0.5\gamma^\star\mathcal L_{\mathrm{SFT}}\) \\
OPD + SFT & Uniform & Direct only & \(\mathcal L_{\mathrm{OPD}}+\gamma^\star\mathcal L_{\mathrm{SFT}}\) \\
OPD + SFT (double) & Uniform & Direct only & \(\mathcal L_{\mathrm{OPD}}+2\gamma^\star\mathcal L_{\mathrm{SFT}}\) \\
MetaOPD + SFT & All 74 descriptors & Outer + direct & \(\mathcal L_{\mathrm{OPD}}^w+\gamma^\star\mathcal L_{\mathrm{SFT}}\) \\
\bottomrule
\end{tabular}
\end{table}

\textbf{Feature-ablation controls and results.}
\Cref{tab:metaopd-feature-ablation} compares uniform OPD, four restricted-input weighting networks, and MetaOPD with all descriptors. The Uniform OPD and MetaOPD rows reuse the Qwen3-1.7B-Base results from \cref{tab:main-results}.

The restricted variants retain, respectively, normalized position, vocabulary-normalized teacher entropy, the teacher-minus-student log-probability gap \(\delta_{i,t}\), or the two ranked top-32 log-probability profiles. All other descriptor coordinates are zeroed while the 74-dimensional input and 74--256--64--1 architecture remain fixed. All learned variants have 35,713 parameters and follow the shared training protocol with seed 42, the shared virtual-update rule, and 696 student updates. This design compares named input subsets with the full descriptor rather than removing one feature at a time.

\textbf{Direct-supervision ablations.}
\Cref{tab:metaopd-gold-ablation} reports SFT, the SFT loss-weight sweep under uniform OPD, and the interaction control that adds the SFT loss to MetaOPD. The Uniform OPD and MetaOPD rows reuse the 1.7B main-table results. All controls use the shared EOS convention, training seed 42, and 696 student updates. The MetaOPD rows retain the shared virtual-update rule; SFT and the uniform-OPD sweep have no meta-update.

\begin{table*}[t]
\centering
\caption{Feature-composition ablation on Qwen3-1.7B-Base (\%).}
\label{tab:metaopd-feature-ablation}
\resizebox{\textwidth}{!}{%
\begin{tabular}{lrrrrrrrrrrrrrr}
\toprule
Variant & \multicolumn{7}{c}{Avg@8} & \multicolumn{7}{c}{Pass@8} \\
\cmidrule(lr){2-8}\cmidrule(lr){9-15}
 & MATH500 & AMC23 & Minerva & HMMT & AIME24 & AIME25 & Avg. & MATH500 & AMC23 & Minerva & HMMT & AIME24 & AIME25 & Avg. \\
\midrule
Uniform OPD & 67.23 & 38.75 & \underline{27.80} & 1.67 & 8.33 & 6.67 & 25.07 & 84.20 & 67.50 & \underline{47.43} & \underline{6.67} & 20.00 & 16.67 & 40.41 \\
Position only & 68.50 & 39.69 & 26.01 & 1.67 & 9.58 & 5.83 & 25.21 & 88.00 & 67.50 & 43.75 & 3.33 & 20.00 & 13.33 & 39.32 \\
Teacher entropy only & 68.88 & \textbf{43.75} & \textbf{27.85} & \textbf{2.50} & \underline{11.25} & 6.25 & \underline{26.75} & 86.60 & 72.50 & 47.06 & \textbf{10.00} & \underline{26.67} & 16.67 & 43.25 \\
Log-probability gap only & 68.50 & 39.38 & 26.75 & 1.67 & 8.75 & \underline{7.50} & 25.42 & 87.80 & \underline{75.00} & 45.59 & \underline{6.67} & \underline{26.67} & \underline{20.00} & \underline{43.62} \\
Top-32 profiles only & \underline{69.05} & \underline{40.94} & 27.57 & 1.67 & 8.33 & 6.67 & 25.70 & \textbf{88.40} & \underline{75.00} & 46.69 & 3.33 & 20.00 & \underline{20.00} & 42.24 \\
MetaOPD (all features) & \textbf{69.95} & \textbf{43.75} & 27.71 & \underline{2.08} & \textbf{12.08} & \textbf{8.33} & \textbf{27.32} & \underline{88.20} & \textbf{77.50} & \textbf{48.53} & \textbf{10.00} & \textbf{33.33} & \textbf{23.33} & \textbf{46.82} \\
\bottomrule
\end{tabular}%
}
\vspace{1mm}
\parbox{0.98\textwidth}{\footnotesize All runs use the same six-benchmark evaluation protocol with eight samples per prompt. Avg. denotes the unweighted average. Bold and underline mark the highest and second-highest distinct scores in each column, including ties. Restricted variants retain only the named descriptor inputs; top-32 profiles concatenate the student's and teacher's ranked log-probability profiles.}
\end{table*}


\begin{table*}[t]
\centering
\caption{SFT and combined OPD--SFT controls on Qwen3-1.7B-Base (\%).}
\label{tab:metaopd-gold-ablation}
\resizebox{\textwidth}{!}{%
\begin{tabular}{lrrrrrrrrrrrrrr}
\toprule
Variant & \multicolumn{7}{c}{Avg@8} & \multicolumn{7}{c}{Pass@8} \\
\cmidrule(lr){2-8}\cmidrule(lr){9-15}
 & MATH500 & AMC23 & Minerva & HMMT & AIME24 & AIME25 & Avg. & MATH500 & AMC23 & Minerva & HMMT & AIME24 & AIME25 & Avg. \\
\midrule
Uniform OPD & 67.23 & \underline{38.75} & \textbf{27.80} & 1.67 & \underline{8.33} & \underline{6.67} & \underline{25.07} & 84.20 & 67.50 & 47.43 & \underline{6.67} & 20.00 & 16.67 & 40.41 \\
SFT & 32.85 & 14.06 & 11.76 & 1.25 & 0.83 & 0.00 & 10.13 & 65.40 & 52.50 & 34.93 & \underline{6.67} & 6.67 & 0.00 & 27.69 \\
OPD + SFT (\(0.5\gamma^\star\)) & \underline{67.45} & 38.44 & 23.99 & \textbf{2.50} & 7.92 & \underline{6.67} & 24.49 & 86.60 & \underline{75.00} & 43.75 & \textbf{10.00} & 16.67 & 16.67 & 41.45 \\
OPD + SFT (\(\gamma^\star\)) & 63.00 & 33.75 & 23.53 & \underline{2.08} & 7.50 & 6.25 & 22.69 & 86.20 & \textbf{77.50} & 45.59 & \underline{6.67} & \underline{23.33} & 13.33 & \underline{42.10} \\
OPD + SFT (\(2\gamma^\star\)) & 57.25 & 27.81 & 22.75 & 1.25 & \underline{8.33} & 5.42 & 20.47 & 86.20 & 67.50 & \underline{47.79} & \textbf{10.00} & 20.00 & \underline{20.00} & 41.92 \\
MetaOPD & \textbf{69.95} & \textbf{43.75} & \underline{27.71} & \underline{2.08} & \textbf{12.08} & \textbf{8.33} & \textbf{27.32} & \textbf{88.20} & \textbf{77.50} & \textbf{48.53} & \textbf{10.00} & \textbf{33.33} & \textbf{23.33} & \textbf{46.82} \\
MetaOPD + SFT & 60.20 & 28.44 & 23.81 & \underline{2.08} & 7.92 & 6.25 & 21.45 & \underline{86.80} & 57.50 & 46.32 & \underline{6.67} & 20.00 & \textbf{23.33} & 40.10 \\
\bottomrule
\end{tabular}%
}
\vspace{1mm}
\parbox{0.98\textwidth}{\footnotesize All runs use the same six-benchmark evaluation protocol with eight samples per prompt. Avg. denotes the unweighted average; benchmark and Avg. scores are rounded separately. Bold and underline mark the highest and second-highest distinct scores in each column, including ties. SFT trains the student directly on reference solutions; OPD + SFT and MetaOPD + SFT add the SFT loss to the student objective at the indicated coefficient, with \(\gamma^\star\approx2.16\).}
\end{table*}


\textbf{Resource accounting.}
\Cref{tab:metaopd-resource} compares OPD and MetaOPD with the 1.7B student, training seed 42, and the same eight-GPU allocation. Both runs use 174 rollout iterations, rollout batches of 128, optimizer mini-batches of 32, and 696 student updates. MetaOPD performs a meta-update at each of the 696 student updates; OPD has no outer objective. The MetaOPD timing uses response-centered scores and the coupled schedules in Appendix~\ref{app:meta-schedule}.

Training wall time is multiplied by all eight allocated GPUs, including the four paired teacher GPUs. Evaluation and diagnostic probes are excluded. The resulting 45.0 and 88.5 GPU-hours give a \(1.97\times\) ratio for this 1.7B setting. These measurements quantify the cost of the complete training procedure, including the outer objective, rather than the small weighting network alone.

\begin{table}[htbp]
\centering
\small
\caption{Training-only cost for the 1.7B student on eight NVIDIA A100 80GB GPUs, including four student and four teacher GPUs. Both methods complete 696 student updates; MetaOPD performs one meta update per step. Evaluation and diagnostic probes are excluded.}
\label{tab:metaopd-resource}
\begin{tabular}{@{}lrrr@{}}
\toprule
Method & Training time (h) & GPU-hours & Relative cost \\
\midrule
OPD & 5.62 & 45.0 & \(1.00\times\) \\
MetaOPD & 11.06 & 88.5 & \(1.97\times\) \\
\bottomrule
\end{tabular}
\end{table}


\section{Motivating One-Step Comparison}
\label{app:intro-motivation-diagnostic}

\subsection{Shared Setting}

The comparison in \cref{fig:intro-motivation} uses Qwen3-1.7B-Base as the student and Qwen3-8B as the frozen teacher. Within each panel, all strategies start from the same initial student and optimizer state and perform one actual student update on the same student-generated responses. The optimizer starts with zero AdamW moments and completed-step count zero, uses a learning rate of \(3\times10^{-6}\), and clips the global gradient norm at 1.0. Virtual and actual student updates use the same AdamW update rule, with the virtual update leaving the shared starting state unchanged. The completion mask retains the valid sampled EOS action.

We use four fixed prompt folds and three response-sampling seeds, giving 12 panels. Each panel contains 16 prompts with two responses per prompt. The seeds change only response sampling; the prompt selection and fold assignment remain fixed across seeds. We also sample two independent reference sets, A and B, from the MATH training pool, each containing 64 prompts and their reference solutions, and share both sets across all panels.

\subsection{Compared Strategies}

Uniform OPD assigns equal weights to valid response tokens. MetaOPD starts from a weighting network initialized to produce uniform weights. It uses reference set A to compute the validation loss after a virtual weighted OPD update, performs one meta-update of the network, and recomputes the token weights. These weights then determine the actual student update from the original student parameters and optimizer state. Reference set B evaluates the resulting student and does not enter weight learning.

EOPD~\citep{jin2026entropy} and TIP~\citep{xu2026tip} apply their supervision-allocation rules within the shared sampled-OPD framework. TIP uses the Soft-OR selection described in Appendix~\ref{app:token-allocation-baselines}. Random assigns random nonuniform token weights, providing a reference for allocation without validation-loss feedback. All candidate student updates use the shared responses, learning rate, and clipping threshold described above.

\subsection{Evaluation and Aggregation}

Let \(\theta'_h\) be the student after the actual update under strategy \(h\). We evaluate its teacher-forced reference-solution loss on set B, averaged over valid reference tokens, and report the paired reduction relative to Uniform OPD:
\begin{equation}
    R_h
    =\mathcal L_{\mathrm{out}}(\theta'_{\mathrm{Uniform}};B)
     -\mathcal L_{\mathrm{out}}(\theta'_h;B).
\label{eq:intro-loss-reduction}
\end{equation}
Positive \(R_h\) means that strategy \(h\) yields a lower post-update validation loss than Uniform OPD, matching the direction in \cref{fig:intro-motivation-b}. We average the paired reductions across the 12 panels and obtain 95\% percentile confidence intervals from 10,000 bootstrap resamples of panels, retaining the pairing between strategies. These intervals are conditional on the fixed prompts and reference sets; the response-sampling seeds are not independent full training runs.

In this comparison, MetaOPD achieves the largest observed reduction, EOPD yields a smaller improvement, and TIP and Random remain close to Uniform OPD. The result supports using post-update performance to learn token weights in this one-step setting; it does not establish the effects of repeatedly applying the strategies over a full training run.

\section{Training Dynamics and Update Analysis}
\label{app:checkpoint-credit-diagnostic}

\subsection{Diagnostic Setting}

\begin{figure*}[t]
    \centering
    \subfloat[Weight dispersion\label{fig:training-dynamics-a}]{%
        \includegraphics[width=0.32\textwidth,trim=0pt 13pt 360pt 26.5pt,clip]{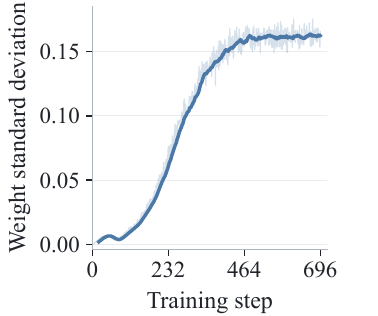}%
    }\hfill
    \subfloat[Weight alignment\label{fig:training-dynamics-b}]{%
        \includegraphics[width=0.32\textwidth,trim=180pt 13pt 180pt 26.5pt,clip]{training_dynamics.pdf}%
    }\hfill
    \subfloat[Update quality\label{fig:training-dynamics-c}]{%
        \includegraphics[width=0.32\textwidth,trim=360pt 13pt 0pt 26.5pt,clip]{training_dynamics.pdf}%
    }
    \caption{\textbf{Training dynamics and update quality (1.7B).}
    (a) Token-weight standard deviation over training.
    (b) Median within-response Spearman correlation between pre-update token weights and \(c_{i,t}\), the scaled negative validation-loss derivative, in epoch 1 of a matched diagnostic run; shading shows the interquartile range.
    (c) Paired validation loss difference \(\Delta L_{\mathrm{Learned}-\mathrm{Uniform}}\) after one update from identical starting parameters at each of three checkpoints; negative values favor Learned. Error bars show 95\% reference-bank bootstrap intervals with checkpoints and rollouts fixed.
    Appendix~\ref{app:checkpoint-credit-diagnostic} gives the full protocol.}
    \label{fig:training-dynamics}
\end{figure*}

We use online traces to track token-weight evolution and fixed-checkpoint probes to examine agreement of loss derivatives across batches, learned allocation, and one-step update quality. Here \(c_{i,t}\) is proportional to the negative validation-loss derivative with respect to a token weight, as defined in \cref{eq:weight-hypergradient}. \Cref{fig:training-dynamics-a} shows the weight standard deviation recorded during training. \Cref{fig:training-dynamics-b} uses a separate, matched diagnostic reproduction with the full three-epoch, 696-step learning-rate schedule, recording only steps 1--232. It reports the median and interquartile range (IQR) of within-response Spearman correlations between pre-update token weights and \(c_{i,t}\) computed using the current training reference batch. Light traces show per-step statistics; dark curves use complete trailing 20-step and 8-step arithmetic means in \cref{fig:training-dynamics-a,fig:training-dynamics-b}, respectively. Constant initial weights make correlation undefined, so that point is left blank.

The checkpoint probes examine Qwen3-1.7B-Base trained with MetaOPD and the restricted-input variants of \cref{sec:ablations}. Early, middle, and final denote MetaOPD steps 232, 464, and 696. The probes use each checkpoint's student AdamW state, with global gradient-clipping threshold 1.0. All candidates use the same AdamW update rule as training at reference learning rate \(3\times10^{-6}\), independent of the decayed training rate, and are not committed to the model.

For the checkpoint probes, student-generated responses provide the data for candidate OPD updates, while reference solutions are used to compute validation loss. Checkpoints share the same rollout prompts but generate responses separately. Each checkpoint's panel contains 64 responses, divided into two groups of 32; candidate updates are computed separately for each group from that checkpoint. All candidate weighting rules and reference banks at a given checkpoint reuse these responses. We also use 256 separate prompt--reference-solution pairs, partitioned into 32 reference banks of eight examples each. The rollout prompts and reference examples are drawn from MATH test with MATH500 excluded. Within each reference bank, validation loss is averaged over valid reference-solution tokens.

We compute the raw \(c_{i,t}\) at the weights produced by the checkpoint's weighting network and define the centered diagnostic quantity
\begin{equation}
 c^\circ_{i,t}
 =c_{i,t}-\frac{1}{n_i}\sum_s m_{i,s}c_{i,s},
 \label{eq:checkpoint-centered-credit}
\end{equation}
where \(n_i=\sum_s m_{i,s}>0\), using the ordinary mean over valid tokens within each response. The raw \(c_{i,t}\) retains the derivative meaning in \cref{eq:weight-hypergradient}; \(c^\circ_{i,t}\) is a centered statistic, distinct from the weighted centering term in the derivative with respect to unnormalized weights. The diagnostic sign and magnitude comparisons below use \(c^\circ_{i,t}\) and \(\lvert c^\circ_{i,t}\rvert\). Subtracting a response-wise constant does not change within-response Spearman correlations or ranks.

To assess consistency across reference batches, we combine the 32 banks into eight disjoint batches of 32 reference examples. For each batch, we average the vectors \(\boldsymbol d\) from its four banks, weighted by their valid reference-token counts, then compute and center \(c_{i,t}\). We compare the resulting \(c^\circ_{i,t}\) values for all pairs of batches within each response using Spearman correlation, sign agreement, and top-20\% Jaccard overlap, then take the median over pairs and responses.

To assess learned-weight alignment, we average \(\boldsymbol d\) across all 32 reference banks, weighted by their valid reference-token counts, using all 256 reference examples. We report the median within-response correlations of learned weights with the resulting \(c^\circ_{i,t}\) values or their magnitudes. The 95\% percentile intervals use 2,000 prompt-cluster bootstrap resamples, retaining all observations for each sampled prompt. This token-count weighting defines the derivative aggregation; the one-step loss comparisons below instead average banks equally.

\subsection{Learned Weights and Update Quality}
\label{app:trained-credit}
\label{sec:credit-analysis}

In the 1.7B setting, weight dispersion increases from uniform initialization and then levels off (\cref{fig:training-dynamics-a}), while each response's coefficient sum remains fixed. In the matched diagnostic run, the online correlation between token weights and \(c_{i,t}\) turns from negative to positive and reaches about 0.3 later in the recorded first epoch (\cref{fig:training-dynamics-b}).

The online curve uses the current training reference batch, whereas checkpoint alignment uses that checkpoint's rollout panel and \(c^\circ_{i,t}\) computed from the averaged vector over 256 reference examples, so their correlations are not interchangeable.

The \(c^\circ_{i,t}\) values agree across independent reference batches at all three checkpoints (rank correlation around 0.98; sign agreement above 0.99; \cref{tab:trained-credit}a), providing a reproducible basis for analyzing the learned weights.

Learned weights correlate positively with \(c^\circ_{i,t}\) at all three checkpoints, with 95\% bootstrap intervals above zero (\cref{tab:trained-credit}b); the final correlation is 0.267 \([0.246,0.278]\). The intervals for correlations with \(\lvert c^\circ_{i,t}\rvert\) include zero, leaving that association inconclusive. Because the student and optimizer states change, the derivatives used for comparison also change; these correlations therefore do not establish convergence of the weighting rule.

\begin{table}[htbp]
\centering
\small
\caption{Loss-derivative agreement and learned-weight correlations for MetaOPD at three training stages (1.7B). Agreement compares \(c^\circ_{i,t}\) computed from 32-example reference batches; weight correlations use \(c^\circ_{i,t}\) obtained after averaging \(\boldsymbol d\) over 256 reference examples, weighted by valid reference-token counts, and centering within each response. Values summarize within-response comparisons. Brackets report 95\% intervals from 2,000 prompt-cluster bootstrap resamples.}
\label{tab:trained-credit}
\begin{tabular}{@{}lrrrr@{}}
\toprule
\multicolumn{5}{l}{\textit{(a) Agreement across independent reference batches}} \\
Checkpoint & Student steps & Spearman & Sign agreement & Top-20\% Jaccard \\
\midrule
Early & 232 & 0.975 & 0.993 & 0.958 \\
Middle & 464 & 0.976 & 0.993 & 0.961 \\
Final & 696 & 0.975 & 0.993 & 0.953 \\
\midrule
\multicolumn{5}{l}{\textit{(b) Weight alignment: Spearman correlation [95\% CI]}} \\
Checkpoint & \multicolumn{2}{c}{\(c^\circ_{i,t}\)} & \multicolumn{2}{c}{\(\lvert c^\circ_{i,t}\rvert\)} \\
\midrule
Early & \multicolumn{2}{c}{\(0.271\;[0.249, 0.286]\)} & \multicolumn{2}{c}{\(0.004\;[-0.116, 0.169]\)} \\
Middle & \multicolumn{2}{c}{\(0.266\;[0.255, 0.284]\)} & \multicolumn{2}{c}{\(0.081\;[-0.093, 0.190]\)} \\
Final & \multicolumn{2}{c}{\(0.267\;[0.246, 0.278]\)} & \multicolumn{2}{c}{\(0.082\;[-0.064, 0.204]\)} \\
\bottomrule
\end{tabular}
\end{table}


At the final checkpoint, correlations with \(c^\circ_{i,t}\) are 0.007 for position, 0.216 for teacher entropy, and \(-0.569\) for the teacher--student log-probability gap. The gap has a larger absolute correlation than the learned weights (0.267), in the opposite direction. These within-response correlations do not directly measure validation loss after a student update.

\subsection{One-Step Update Comparisons}
We compare Uniform, Learned, and Reweighted allocations from the same checkpoint, computing candidate updates separately for each of the two rollout groups. Learned uses weights directly from the checkpoint's frozen weighting network, without a new meta-update; this probe evaluates the allocation already learned at that checkpoint. Reweighted multiplies these learned weights by a linear factor in \([0.9,1.1]\) based on the within-response rank of \(c^\circ_{i,t}\), then restores unit response mean. It uses the same eight-example reference bank to compute the ranking and evaluate the update, so this comparison gives Reweighted access to its evaluation objective.

Let \(B_b\) denote reference bank \(b\) and \(\theta'_{h,b,g}\) the candidate for strategy \(h\), bank \(b\), and rollout group \(g\). Within each bank, we compute token-mean validation loss and average the two group results equally, then average the 32 banks equally:
\begin{equation}
 \Delta L_h^{\mathrm{step}}
 =\frac{1}{32}\sum_{b=1}^{32}\frac{1}{2}\sum_{g=1}^{2}
 \left[\mathcal L_{\mathrm{out}}(\theta'_{h,b,g};B_b)
       -\mathcal L_{\mathrm{out}}(\theta;B_b)\right],
 \label{eq:trained-absolute-intervention}
\end{equation}
where \(h\in\{\mathrm{Uniform},\mathrm{Learned},\mathrm{Reweighted}\}\). Negative \(\Delta L_h^{\mathrm{step}}\) indicates a lower mean loss than the unchanged checkpoint. Uniform and Learned do not use the bank to construct their weights, so their candidates depend only on the rollout group. The paired contrasts are
\begin{align}
 \Delta L_{\mathrm{Learned}-\mathrm{Uniform}}
 &=\Delta L_{\mathrm{Learned}}^{\mathrm{step}}
   -\Delta L_{\mathrm{Uniform}}^{\mathrm{step}},
 \label{eq:checkpoint-intervention-difference}\\
 \Delta L_{\mathrm{Reweighted}-\mathrm{Learned}}
 &=\Delta L_{\mathrm{Reweighted}}^{\mathrm{step}}
   -\Delta L_{\mathrm{Learned}}^{\mathrm{step}}.
 \label{eq:checkpoint-reweighted-difference}
\end{align}
The pre-update losses cancel because each contrast uses the same banks and rollout groups.

For Learned--Uniform (\cref{fig:training-dynamics-c}; \cref{tab:trained-intervention}), we average the two paired group differences within each bank, then resample the 32 banks with replacement 10,000 times and recompute their equally weighted mean to obtain a 95\% percentile interval. These intervals condition on the checkpoint and its fixed responses, excluding training-seed variation.

At the diagnostic reference learning rate of \(3\times10^{-6}\), Learned yields lower validation loss than Uniform at all three checkpoints (\cref{fig:training-dynamics-c,tab:trained-intervention}), with a final paired difference of approximately \(-1.95\times10^{-3}\). Both allocations nevertheless increase mean loss relative to the unchanged checkpoint at every stage, so Learned's advantage is relative rather than absolute.

Reweighted yields lower validation loss than Learned at each checkpoint, although all three allocations increase loss at the final checkpoint (\cref{tab:trained-intervention}). These interventions use local derivative rankings to take finite steps at the reference rate; their relative ordering does not guarantee descent or improvement over the full training trajectory.

\begin{table}[htbp]
\centering
\small
\setlength{\tabcolsep}{3pt}
\caption{One-step validation loss changes \(\Delta L_h^{\mathrm{step}}\) and paired contrasts (\(10^3\) scale; lower is better) at reference learning rate \(3\times10^{-6}\). Learned$-$Uniform intervals use 10,000 reference-bank bootstrap resamples with checkpoint and rollouts fixed. Reweighted uses the evaluation bank to construct weights.}
\label{tab:trained-intervention}
\begin{tabular*}{\textwidth}{@{\extracolsep{\fill}}lrrrrrr@{}}
\toprule
\multirow{2}{*}{\shortstack[l]{Checkpoint\\(step)}}
& \multicolumn{3}{c}{\textit{(a) Validation loss changes}}
& \multicolumn{3}{c}{\textit{(b) Paired contrasts}} \\
\cmidrule(lr){2-4}\cmidrule(l){5-7}
& Uniform & Learned & Reweighted & Learned$-$Uniform & 95\% CI & Reweighted$-$Learned \\
\midrule
Early (232) & \(+0.74\) & \(+0.09\) & \(-3.32\) & \(-0.65\) & \([-0.72, -0.59]\) & \(-3.41\) \\
Middle (464) & \(+2.78\) & \(+0.80\) & \(-3.03\) & \(-1.97\) & \([-2.16, -1.80]\) & \(-3.83\) \\
Final (696) & \(+6.26\) & \(+4.31\) & \(+0.61\) & \(-1.95\) & \([-2.13, -1.78]\) & \(-3.70\) \\
\bottomrule
\end{tabular*}
\end{table}


\textbf{Restricted-input allocations.}
All four restricted-input variants are evaluated at student step 696 and yield consistent derivative estimates and negative Reweighted--Learned
contrasts under the same protocol (\cref{tab:checkpoint-credit}), assessing local update
quality rather than benchmark accuracy.

\begin{table}[!htb]
\centering
\small
\caption{Loss-derivative agreement and update comparisons for restricted-input variants (1.7B), all at student step 696. Agreement uses \(c^\circ_{i,t}\) computed from 32-example reference batches; \(\Delta L_{\mathrm{Reweighted}-\mathrm{Learned}}\) uses equally weighted eight-example reference banks at reference rate \(3\times10^{-6}\). Point estimates; negative contrasts favor Reweighted.}
\label{tab:checkpoint-credit}
\begin{tabular}{@{}lrrrr@{}}
\toprule
Input restriction & Spearman & Sign agreement & Top-20\% Jaccard & \(10^3\Delta L_{\mathrm{Reweighted}-\mathrm{Learned}}\) \\
\midrule
Position only & 0.978 & 0.994 & 0.962 & -3.56 \\
Teacher entropy & 0.978 & 0.994 & 0.961 & -3.92 \\
Log-prob. gap & 0.979 & 0.993 & 0.959 & -3.84 \\
Top-32 profiles & 0.975 & 0.992 & 0.960 & -3.30 \\
\bottomrule
\end{tabular}
\end{table}


\clearpage
\section{Out-of-Domain Generalization}
\label{app:ood-generalization}

We evaluate the Qwen3-1.7B-Base checkpoints from the MATH main experiment on GPQA-Diamond~\citep{rein2023gpqa}, MMLU-Pro~\citep{wang2024mmlu}, and AlpacaEval 2.0~\citep{dubois2024length}, without additional training on these tasks. Distillation methods use the Qwen3-8B teacher. All entries in \cref{tab:ood-results} are obtained from local evaluation under a common protocol for each benchmark; Base Model denotes Qwen3-1.7B-Base before MATH training.

\textbf{Evaluation protocol.}
Across these benchmarks, we use temperature 1.0, top-\(p\) 0.8, and a maximum response length of 8192 tokens, matching the underlying decoding settings in \cref{tab:metaopd-reproducibility}. The number of responses and evaluation metric are benchmark-specific.

GPQA-Diamond contains 198 questions, for which we sample eight responses per question. Avg@8 measures average response accuracy, and Pass@8 measures the fraction of questions answered correctly at least once. MMLU-Pro contains 12,032 test questions and uses five-shot chain-of-thought demonstrations matched to the question's category. We generate one response per question and report accuracy as Pass@1. AlpacaEval 2.0 compares model responses with GPT-4-Turbo reference responses using the \texttt{weighted\_alpaca\_eval\_gpt4\_turbo} annotator. We report both raw win rate (WR) and length-controlled win rate (LC-WR); these are preference-based scores, rather than answer-verification accuracy.

\textbf{Aggregation.}
All scores are on a 0--100 scale. The Avg. row is the unweighted arithmetic average of the five displayed metric scores:
\[
\begin{aligned}
\mathrm{Avg}=\tfrac{1}{5}\bigl(&\mathrm{GPQA}_{\mathrm{Avg@8}}+\mathrm{GPQA}_{\mathrm{Pass@8}}
+\mathrm{MMLU}_{\mathrm{Pass@1}}\\
&+\mathrm{Alpaca}_{\mathrm{LC\text{-}WR}}+\mathrm{Alpaca}_{\mathrm{WR}}\bigr).
\end{aligned}
\]
This descriptive summary assigns 40\%, 20\%, and 40\% of its weight to GPQA-Diamond, MMLU-Pro, and AlpacaEval, respectively. It is distinct from the six-benchmark mathematical-reasoning averages in \cref{tab:main-results}; individual metrics remain necessary to interpret the transfer profile.

\textbf{Results.}
MetaOPD has the highest reported scores on both GPQA-Diamond metrics and MMLU-Pro Pass@1. Its AlpacaEval scores are close to OPD's and below TIP's, showing that the observed advantage varies across the evaluated tasks. MetaOPD also has the highest descriptive average across the five reported metrics.

\end{document}